\documentclass[10pt,twocolumn,letterpaper]{article}

 \usepackage{cvpr}              
\definecolor{cvprblue}{rgb}{0.21,0.49,0.74}
\usepackage[pagebackref,breaklinks,colorlinks,allcolors=cvprblue]{hyperref}

\usepackage{booktabs}
\usepackage{multirow}
\usepackage{tabularx}
\usepackage{enumitem}
\usepackage{ragged2e}
\usepackage{makecell}
\usepackage{xcolor}
\usepackage{hyperref}
\definecolor{AcaRed}{HTML}{8B0000}
\newcolumntype{Y}{>{\raggedright\arraybackslash}X}

\usepackage[most]{tcolorbox}
\newtcblisting{promptbox}{
  breakable,
  colback=gray!12,
  colframe=black,
  boxrule=0.4pt,
  arc=2pt,
  left=6pt,right=6pt,top=6pt,bottom=6pt,
  listing only,
  listing options={basicstyle=\ttfamily\footnotesize,breaklines=true,columns=fullflexible},
  title={System Prompt}
}
\UseRawInputEncoding

\def\paperID{xxx} 
\def\confName{CVPR}
\def\confYear{2026}

\title{A$^3$: Towards Advertising Aesthetic Assessment}

\author{Kaiyuan Ji$^{1,3}$ \quad Yixuan Gao$^{2}$\thanks{Corresponding author.} \quad Lu Sun$^{1,4}$ \quad Yushuo Zheng$^{1,2}$ \quad Zijian Chen$^{1,2}$ \\ \quad Jianbo Zhang$^{1,2}$ \quad Xiangyang Zhu$^{1}$ \quad Yuan Tian$^{1}$ \quad Zicheng Zhang$^{1,2}$ \\ \quad Guangtao Zhai$^{1,2,3}$\footnotemark[1]\\
$^1$Shanghai Artificial Intelligence Laboratory\\
$^2$Institute of Image Communication and Network Engineering, Shanghai Jiao Tong University\\
$^3$School of Information and Electronic Engineering, East China Normal University\\
$^4$School of Computer Science and Technology, Xi'an Jiaotong University
}

\begin{document}
\sloppy 
\setlength{\tabcolsep}{4pt} 
\maketitle
 \begin{abstract}
Advertising images significantly impact commercial conversion rates and brand equity, yet current evaluation methods rely on subjective judgments, lacking scalability, standardized criteria, and interpretability. To address these challenges, we present \textbf{A$^3$ (Advertising Aesthetic Assessment)}, a comprehensive framework encompassing four components: a paradigm (\textbf{A$^3$-Law}), a dataset (\textbf{A$^3$-Dataset}), a multimodal large language model (\textbf{A$^3$-Align}), and a benchmark (\textbf{A$^3$-Bench}). Central to A$^3$ is a theory-driven paradigm, A$^3$-Law, comprising three hierarchical stages: (1) Perceptual Attention, evaluating perceptual image signals for their ability to attract attention; (2) Formal Interest, assessing formal composition of image color and spatial layout in evoking interest; and (3) Desire Impact, measuring desire evocation from images and their persuasive impact. Building on A$^3$-Law, we construct A$^3$-Dataset with 120K instruction-response pairs from 30K advertising images, each richly annotated with multi-dimensional labels and Chain-of-Thought (CoT) rationales. We further develop A$^3$-Align, trained under A$^3$-Law with CoT-guided learning on A$^3$-Dataset. Extensive experiments on A$^3$-Bench demonstrate that A$^3$-Align achieves superior alignment with A$^3$-Law compared to existing models, and this alignment generalizes well to quality advertisement selection and prescriptive advertisement critique, indicating its potential for broader deployment. Dataset, code, and models can be found at: \href{https://github.com/euleryuan/A3-Align}{\textcolor{AcaRed}{https://github.com/euleryuan/A3-Align}}
\
\end{abstract}    
 \section{Introduction}
\label{sec:intro}

\begin{figure}[t]
  \centering
   \includegraphics[width=0.9\linewidth]{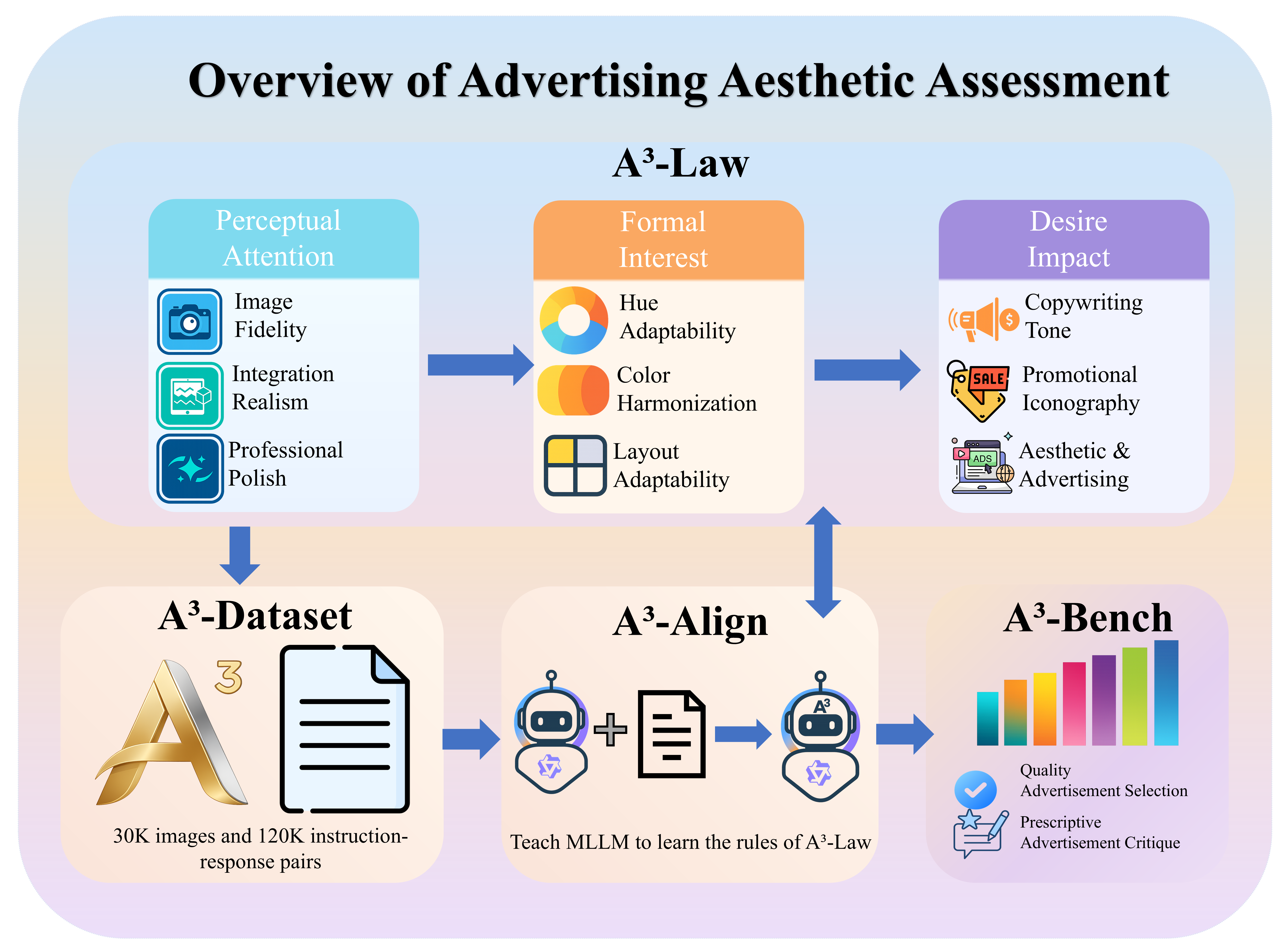}

   \caption{\textbf{Overview of A$^3$: Advertising Aesthetic Assessment.}
A$^3$ centers on the A$^3$-Law, a three stage paradigm with \emph{Perceptual Attention}, \emph{Formal Interest}, and \emph{Desire Impact}. 
Built on this paradigm, \textbf{A$^3$-Dataset} contains 30K images and 120K instruction-response pairs with Chain of Thought; \textbf{A$^3$-Align} learns the rules of A$^3$-Law; and \textbf{A$^3$-Bench} evaluates MLLMs and two tasks such as quality selection and prescriptive critique.}
   \label{fig:figure-1}
\end{figure}

Advertising imagery has become a pervasive part of daily life. However, with the explosive growth of digital media, consumers are now facing severe Advertising Clutter \cite{ha2008integrated,ha2017digital} and Marketing Overload \cite{rehman2023marketing}. This excessive bombardment of advertising messages has led to the dilution of consumer attention, an increase in ad avoidance behaviors, and has demonstrated a significant negative impact on brand recall \cite{rehman2023marketing}. In such a hyper-competitive environment, advertising aesthetics is no longer merely an enhancement; it has become a critical factor for cutting through the noise.

Therefore, while understanding and measuring the aesthetic quality of advertising images is exceptionally important, current evaluation methods still suffer from significant limitations. First, mainstream quality assessment \cite{gao2025blind,gao2025multi,gao2024no,zhang2022no} largely depends on manual subjective scoring \cite{murray2012ava,li2024computer,li2025perceptual,cao2025towards,jiang2026surveillance}, which struggles to achieve consistent consensus and lacks the scalability required to process massive data volumes. Second, existing automated systems \cite{jang2022modeling,dutt2024explainable,ji2024application} often act as mere threshold-based filters \cite{wang2019aspect,jin2025rgcvqa,wang2025learning}, failing to provide the systematic, diagnostic feedback necessary to identify specific shortcomings and guide minor refinements for near-standard advertisements.

In recent years, Multimodal Large Language Models (MLLMs) have demonstrated remarkable general understanding capabilities in image interpretation \cite{yue2024mmmu,liu2024mmbench,chen2025just} and evaluation tasks \cite{guo2025human,zheng2026learningwanderimprovingglobal, zheng2025geoxbench}. However, their current applications in A$^3$ (Advertising Aesthetic Assessment) are largely confined to one-step holistic scoring that neglects progressive human cognition. Furthermore, these models often produce unstable, prompt-sensitive rationales with frequent misalignments between their reasoning \cite{chen2025reasoning,turpin2023language,ji2025medomni,ji2026medomni45} and final outputs \cite{baker2025monitoring,guan2024hallusionbench,ji2025evaluating}. Therefore, there is an urgent need to develop a stepwise evaluation framework that provides suggestions, supported by datasets and training paradigms to ensure reliable and traceable assessments.

To address the theoretical and procedural gaps \cite{si2024accelerating,li2025quantumapproximateoptimizationalgorithms,wang2025bamnet} in A$^3$, we propose a progressive evaluation paradigm named A$^3$-Law, inspired by the AIDA \cite{strong1925theories}. This paradigm deconstructs the A$^3$ into three stages. Perceptual Attention: evaluating perceptual image signals for their ability to attract attention. Formal Interest: assessing the formal composition of image color and spatial layout in evoking interest. Finally, Desire Impact: measuring desire evocation from images and their persuasive impact. The core contribution of $A^3$-Law lies in introducing the first A$^3$ framework that novelly operationalizes abstract theories into an executable hierarchy for annotation, training, and evaluation.

In order to validate the effectiveness of A$^3$-Law and support model training, we construct A$^3$-Dataset, a structured multimodal dataset comprising 120K instruction-response pairs from 30K advertising images with fine-grained labels, Chain-of-Thought (CoT) rationales, and visual annotations. Leveraging this dataset, we train A$^3$-Align via supervised fine-tuning and reinforcement learning to integrate domain knowledge and align evaluation signals. We then develop A$^3$-Bench to evaluate performance, with experiments showing that A$^3$-Align consistently surpasses existing MLLMs. Finally, A$^3$-Align demonstrates strong practical utility in real-world advertisement selection and prescriptive critique by reliably identifying high-quality ads and diagnosing issues with clear explanations.

As shown in Figure \ref{fig:figure-1}, our contributions are as follows:

\begin{itemize}
    \item We propose paradigm \textbf{A$^3$-Law} for automated advertising aesthetics assessment, explicitly decomposing visual evaluation into three theory-driven stages.
    
    \item We construct \textbf{A$^3$-Dataset}, a large-scale dataset containing \textbf{120K} advertising-image annotations aligned with the \textbf{A$^3$-Law}, enabling structured and progressive model training.
    
    \item We establish \textbf{A$^3$-Bench}, a comprehensive benchmark that evaluates numerous mainstream MLLMs.
    \item We demonstrate that \textbf{A$^3$-Law} and \textbf{A$^3$-Align} can enhance two real-world applications: quality advertisement selection and prescriptive advertisement critique.
\end{itemize}


 \section{Related Works}

\textbf{Advertising Aesthetic Assessment.} Current research on advertising images \cite{li2024computer} primarily focuses on attractiveness prediction, click-through rate (CTR) modeling \cite{azimi2012impact,azimi2012visual,zhou2016predicting,bai2025comprehensive}, and aesthetic visual analysis \cite{murray2012ava,talebi2018nima,he2022rethinking,yi2023towards,wang2025learning}, addressing whether ads are effective rather than why. Due to the absence of explicit modeling of aesthetic and persuasive mechanisms \cite{hussain2017automatic,ahuja2018understanding}, crucial factors like copywriting tone \cite{kalra2020understanding,dey2018don,savchenko2020ad}, layout \cite{li2019layoutgan,lee2020neural,zheng2019content}, and color harmony \cite{cohen2006color,hasler2003measuring}remain overlooked, limiting interpretability and generalization \cite{schultze2023explaining,tong2022interpretable,santos2024towards,wang2026brain}. To overcome this, we propose A$^3$-Law, a hierarchical framework for structured, progressive evaluation of advertising aesthetics.

\textbf{Multimodal Large Language Models.} MLLMs offer strong visual-language reasoning \cite{lei2026sequential,jin2025medscreendental} for rule-based evaluation \cite{AIBench,zhang2025large,zhang2025teaching,zhang2025quality}, but without domain alignment they often lack rule awareness \cite{wang2025quality,liang2025priceseer,ji2025assessing,zheng2025lmfightarenabenchmarking} and produce descriptive rather than judgmental outputs \cite{hewitt2024instruction,chen2024internvl,tian2025smc++,tian2025rofi,tian2025towards,wang2026dental}. To address this, we build A$^3$-Dataset directly around the rules of A$^3$-Law and leverage it to train A$^3$-Align.

\textbf{Benchmarks.} Existing benchmarks like AVA \cite{murray2012ava} and AADB \cite{kong2016photo} rely on single-dimensional metrics, overlooking semantic interpretation and inferential quality, and lack frameworks for assessing MLLMs in advertising aesthetics \cite{huang2024aesbench,qi2025photographer,jia2023kafa,tian2024coding,tian2025semantic}. To bridge gap, we propose A$^3$-Bench.

\label{sec:formatting}

\begin{figure*}
  \centering
  \includegraphics[width=1.0\linewidth]{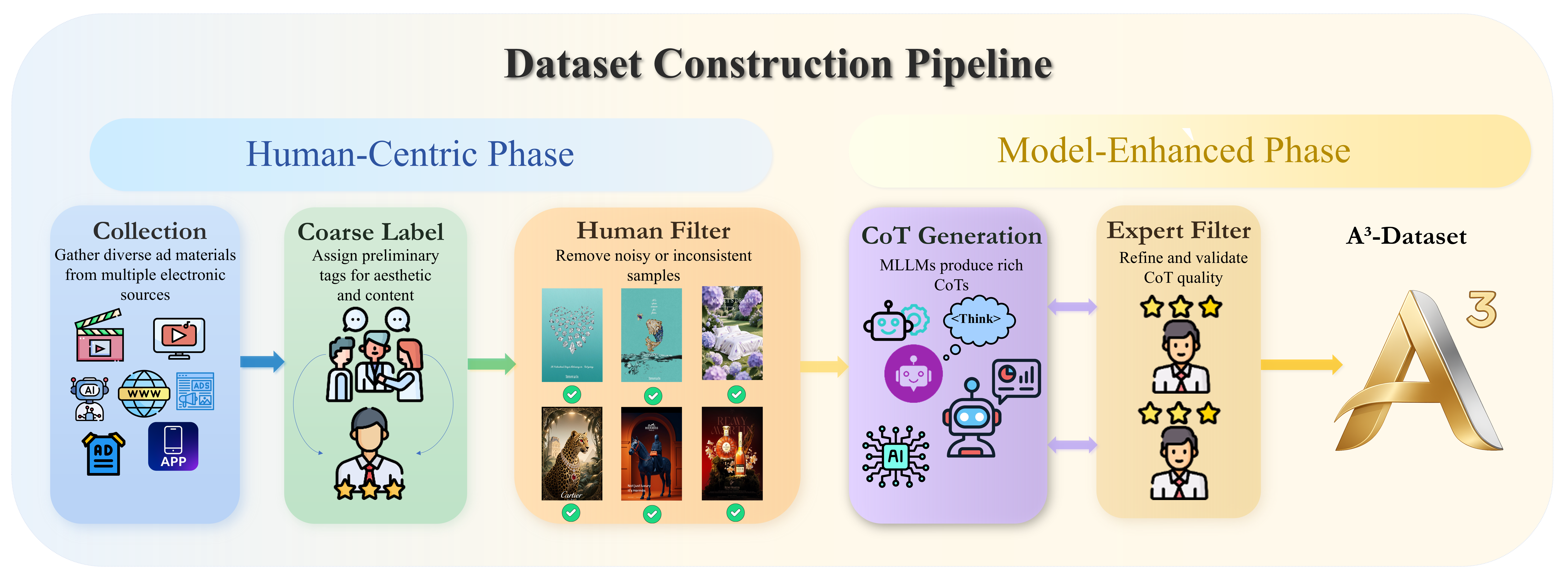}
  \hfill
  \caption{\textbf{A$^3$-Dataset construction pipeline.}
The pipeline has two stages. In the \emph{Human Centric Phase} we collect diverse advertising images, assign preliminary aesthetic and content tags under A$^3$-Law, and remove noisy or inconsistent samples. In the \emph{Model Enhanced Phase} multimodal LLMs generate Chain of Thought rationales that are refined and validated by experts. The result is the A$^3$-Dataset with 30K images and 120K instruction and response pairs of high quality.}
  \label{fig:short}
\end{figure*}

\section{Approach}

\subsection{A$^3$-Law: Hierarchical Paradigm}

\textbf{Perceptual Attention. }It evaluates perceptual image signals for their ability to attract attention, constituting the first physiological threshold for advertising information processing and serving as the unconscious precondition for the ``Attention" stage in the classic AIDA \cite{strong1925theories,barry1987development}model. Rather than performing aesthetic judgment, it functions as an ultra-rapid preprocessing mechanism, completed within a hundred milliseconds \cite{potter2014detecting,thorpe1996speed}, in which the brain must first determine whether the incoming signal is ``processable and valuable information" or should be immediately discarded as ``visual noise." This initial screening is grounded in signal detection theory and information theory \cite{green1966signal,hautus2021detection,shannon1948mathematical}: reliable perceptual admission requires a sufficient signal-to-noise ratio and low distortion, so that the visual input carries enough recoverable information to cross early physiological thresholds. Consequently, we posit the Image Fidelity principle, which states that images should be clear and minimally distorted to ensure effective reception before any higher-level aesthetic judgment can occur.

Second, Perceptual Fluency Theory \cite{johnston1985perceptual,reber1999effects} holds that stimuli that are easier to parse elicit more credible judgments. We therefore operationalize fluency with two design constraints. Integration Realism requires physically coherent rendering, including consistent lighting, color temperature, shadows, and perspective, so that the scene accords with the visual system’s prior expectations and reduces prediction error during early parsing. Professional Polish requires clean, artifact-free textures and legible micro-details that match commercial category prototypes, thereby minimizing processing friction and supporting rapid credibility judgments. Together with Image Fidelity, these constraints form a sequential dependency from input to parsing to trust, enabling signals to pass early physiological thresholds before any higher level aesthetic judgment.

Only by passing this physiological threshold is the advertising image confirmed as a qualified visual signal, thus permitting its entry into higher-order stages.

\begin{figure*}
  \centering
  \includegraphics[width=1.0\linewidth]{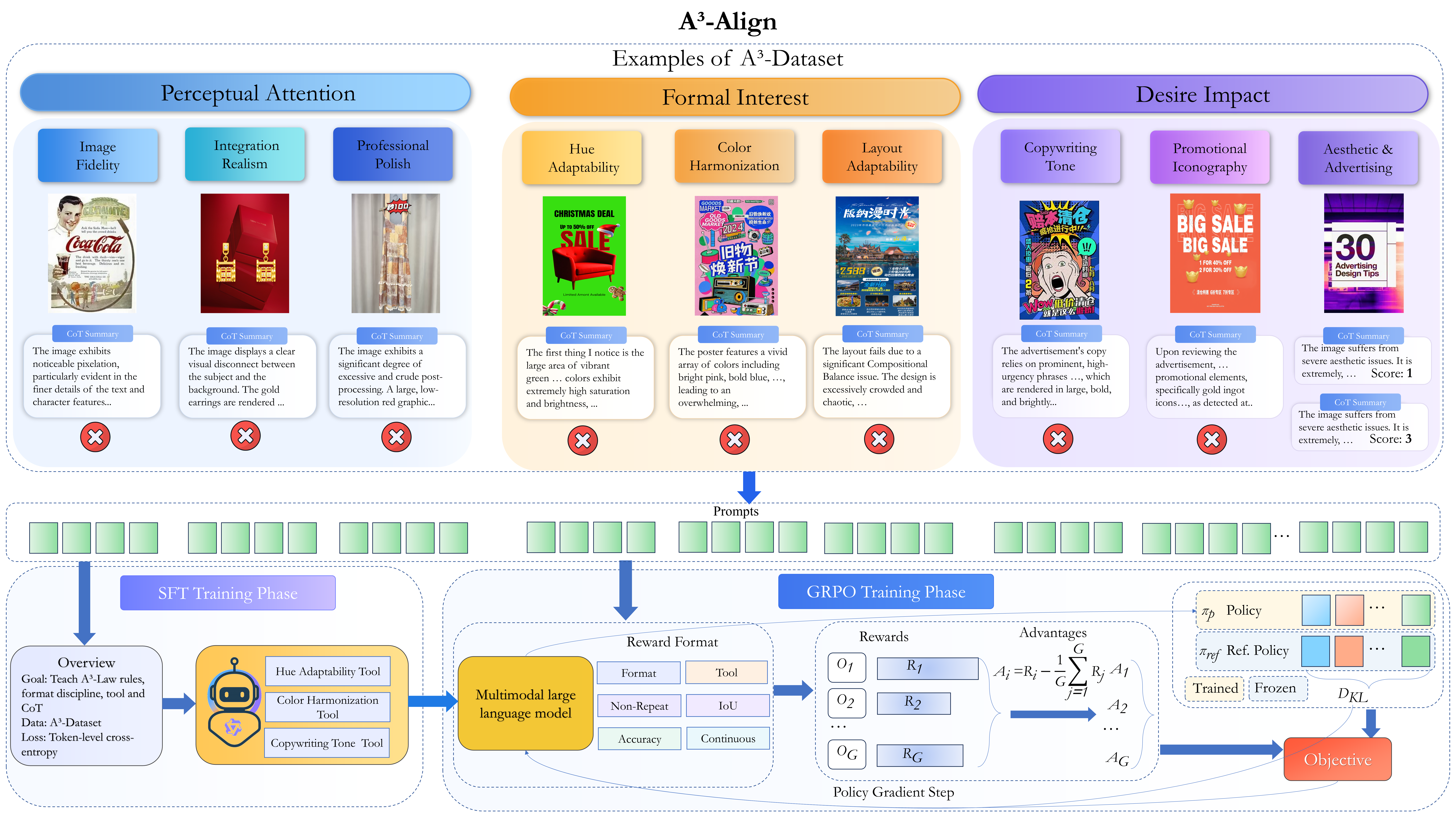}
  \hfill
  \caption{\textbf{A$^3$-Align under the A$^3$-Law.}
The top panel shows examples from the A$^3$-Dataset organized by the three stages \emph{Perceptual Attention}, \emph{Formal Interest}, and \emph{Desire Impact}, with subcriteria and Chain of Thought summaries.
The bottom panel presents a two-phase training pipeline.
In the SFT phase the multimodal LLM learns A$^3$-Law rules, structured output format, tool use, and Chain of Thought from the A$^3$-Dataset with token-level cross-entropy.
In the GRPO phase, the model is optimized with multi-signal rewards, ultimately leading to A$^3$-Align, which produces rule-based judgments.}
  \label{fig:figure-3}
\end{figure*}
\textbf{Formal Interest. }It assesses the formal composition of image color and spatial layout in evoking interest. Following the Perceptual Attention physiological screening, this stage constitutes a higher order cognitive task in which the visual system organizes disparate elements into a meaningful structure. Interest arises when a scene is sufficiently comprehensible yet moderately novel \cite{berlyne1970novelty, silvia2005interesting}. In line with appraisal accounts \cite{silvia2005interesting}, under these conditions, viewers engage a coherence-seeking drive that organizes the scene into a simple, orderly structure. This drive activates fundamental perceptual grouping mechanisms, as described by Gestalt psychology \cite{koffka2013principles}. To execute this grouping task efficiently, the system must prioritize effective organizational cues.

Foundational work on visual perception establishes color as a dominant preattentive feature \cite{treisman1980feature, wolfe2017five}. Therefore, chromatic similarity functions as a primary, highly efficient grouping cue, allowing the system to rapidly carve the scene into coarse, color-based units before refining spatial relations. Anchored in these mechanisms, we formalize Color Construction, whereby similarity in hue, lightness, and saturation promotes stable grouping \cite{cohen2006color}. We operationalize this with two rules: Hue Adaptability, which constrains each hue’s lightness and saturation so that color intensity remains pleasant rather than aversive; and Color Harmonization, which evaluates whether the palette is coherent or instead feels scattered and chaotic. The Hasler colorfulness \cite{hasler2003measuring,ou2006colour} metric serves as a reference for this assessment.

Once the color scheme is coordinated, attention shifts to Spatial Construction \cite{tufte1991envisioning}, which governs how elements are positioned so the design can be parsed quickly and comfortably. Key components (product, text, supporting visuals) are arranged in a clear hierarchy with a single focal point; related items are grouped by proximity, aligned to a simple grid, and spaced consistently. To accommodate aspect-ratio changes and routine platform crops, critical content is placed within safe regions, thereby minimizing information loss when the design is resized or trimmed. The result is a clean, readable layout that avoids clutter. We implement this as Layout Adaptability: a balanced, media-adaptive arrangement that preserves hierarchy, product visibility, and reading order under common format changes.

Together, these rules reduce parsing load and enforce structural clarity, enabling the image to pass coherence threshold and proceed to semantic and affective processing.

\textbf{Desire Impact.} This measures desire evocation from images and their persuasive impact. This stage, corresponding to the ``Desire" stage of the AIDA model, follows the signal screening of the Perceptual Attention  and the structural organization of the Formal Interest. At this stage, cognitive processing shifts from passive parsing to an active evaluation of the image's ``semantic value" and ``affective value," which constitutes the ``value threshold."

Its construction is based on two theoretical pillars. First, Semiotics \cite{barthes1985rhetoric,mick1986consumer} treats the advertisement as a ``visual text'' composed of signs. To assess its ``semantic value,'' we derive the Copywriting Tone rule, which evaluates key ``textual signs,'' and the Promotional Iconography rule, which evaluates ``non-textual signs'' such as promotional icons.

Second, in line with affective design \cite{norman2002emotion,fishwick2004emotional} and Appraisal Theory \cite{scherer2001appraisal} of Emotion, decision propensity at this stage reflects integrated affect rather than structure alone. These accounts distinguish visceral impressions from reflective, meaning-based appraisals. Building on this distinction, it assesses affective value through an overall subjective evaluation rule with two components: Aesthetic Attribute, which quantifies visceral visual pleasure induced by formal features; and Advertising Attribute, which quantifies reflective, brand-anchored emotional connection and persuasion expectancy. Together, these components indicate whether the stimulus crosses the value threshold, complementing the semantic evaluation and advancing the AIDA “Desire” stage.

These three rules jointly evaluate whether an image can effectively convert ``interest" into ``desire". Desire Impact focuses on the clarity of persuasive signals which is a universal commercial prerequisite rather than the localized cultural symbols. Thus, we position $A^3$-Law as a framework, leaving specific cultural calibration for future work.

\subsection{A$^3$-Dataset: Dataset Collection}

Our dataset consists of three main components: images, Chain-of-Thought reasoning processes, and final decisions for each image. To systematically construct A$^3$-Dataset, we designed a rigorous workflow for the entire data collection process. First, we divided the rule-based QA pairs into three categories: (1) binary questions determining whether an image is suitable or unsuitable under a given rule; (2) object detection annotations specific to Promotional Iconography; and (3) subjective ratings for aesthetic and advertising attributes. Based on expert-defined standards, we trained human annotators for image labeling. After each annotation batch, we randomly sampled a portion of the data for quality inspection, comparing the annotations with our gold-standard reference. Annotators were trained accordingly to evaluate images on a 1 to 5 rating scale. For objective metrics, only results with accuracy above 0.93 were accepted; for detection annotations, only batches with average IoU above 0.92 were retained; for subjective ratings, the acceptance threshold was a SRCC above 0.85. According to the specific rule descriptions, annotators provided binary answers, detection results, and rating scores.

After human-aligned annotation, we utilized multiple MLLMs to generate logically consistent CoTs, combining the rule-specific content with the answers. These generated CoTs were then evaluated in batches, with a subset reviewed by a panel of 5 human experts to determine final acceptance. We implemented a two-tiered validation process where, first, an individual CoT process was considered 'accepted' only if it received a majority vote (at least 3 out of 5) from the expert panel. Second, the MLLM generation and review process was iterated until the overall acceptance rate of the evaluated subset—meaning the proportion of CoTs meeting the majority vote—consistently exceeded 85\%.

To enhance reasoning reliability, we introduce a tool-calling subset where MLLMs can access three lightweight analytical tools corresponding to rules with well-defined computational proxies. Specifically, a Hue Analysis Tool computes the central hue, lightness, and saturation of each hue cluster to assist judgments of Hue Adaptability; a Color Harmonization Tool provides the Hasler \cite{hasler2003measuring} colorfulness index as a quantitative reference for palette coherence; and DeepSeek-OCR \cite{wei2025deepseek} extracts textual content for assessing Copywriting Tone. These three rules are selected because their perceptual judgments can be meaningfully informed by quantitative cues, while other rules lack reliable automatic surrogates. Notably, we exclude external object detection for Promotional Iconography, encouraging models to recognize and interpret non-textual promotional symbols autonomously rather than outsourcing perception. In this subset, tool outputs serve as auxiliary evidence integrated into the model’s reasoning, ensuring decisions remain grounded yet not mechanically determined by the tools. The whole process of our pipeline is displayed in Figure \ref{fig:figure-3}.

To ensure robust generalization and prevent domain biases, the curated $A^3$-Dataset maximizes diversity in both commercial categories (e.g., electronics, cosmetics) for universal aesthetic learning, and data sources (e-commerce, social media, web banners), as detailed in the Supplementary.

\subsection{A$^3$-Align: MLLM under A$^3$-Law}

To align the MLLM’s reasoning with the aesthetic framework of A$^3$-Law and endow it with autonomous tool use, we adopt Supervised Fine-Tuning (SFT) and Group Relative Policy Optimization (GRPO) \cite{shao2024deepseekmath}. SFT teaches the model to produce parseable outputs with a CoT and an answer. In the subsequent phase, we introduce a multi-source reward framework that jointly calibrates behavioral form, task accuracy, evidential grounding, and subjective value alignment. The rewards are divided into two categories:

\textbf{General reward:}
(1) Format Reward. It enforces valid tag generation and serves as the foundation for all subsequent rewards:
$
R_{\text{format}} = \mathbf{1}_{\{\text{tags valid}\}}
$

(2) Non-Repeat Reward. To inhibit repetition in the MLLM, we combine a sentence-level term and an n-gram term  with equal weights. First, $
R_{\text{sent}} = 1 - \frac{d}{N}$, where $N$ is the total number of sentences, and $d$ is the number of duplicate sentences.Then, $R_{\text{n-gram}} = \frac{|\text{uniq(n-Grams)}|}{|\text{n-Grams}|}$,
where $|\text{n-Grams}|$ is the total count of n-grams in the text, and $|\text{uniq(n-Grams)}|$ is the count of unique n-grams. Finally, we define this Reward as: 
$R_{\text{nonrep}} = \frac{1}{2}(R_{\text{sent}} + R_{\text{n-gram}})$

These rewards target structural stability: $R_{format}$ ensures parseability and $R_{nonrep}$ prevents degenerative loops, guaranteeing the logical consistency of CoTs.

\textbf{Rule-Specific Alignment Reward: }(1) Accuracy Reward.
This reward is applied to the eight binary rules under A$^3$-Law 
(\textit{Image Fidelity}, \textit{Integration Realism}, \textit{Professional Polish}, 
\textit{Hue Adaptability}, \textit{Color Harmonization}, 
\textit{Layout Adaptability}, \textit{Copywriting Tone}, 
\textit{Promotional Iconography}). 
It evaluates whether the model's predicted label (\(\hat{y}\)) matches the annotated ground truth (\(y\)):
$R_{\text{acc}} = 1\{\hat{y} = y\},
$

(2) Tool Utilization Reward. This reward encourages tool-augmented reasoning within CoT:
This reward is applied to the three tool-assisted rules — Hue Adaptability, Color Harmonization, and Copywriting Tone.
$
R_{\text{tool}} =
1{\text{tool is invoked and referenced in reasoning}}
$

(3) IoU Reward.
This reward is designed for Promotional Iconography. Given a set of \(N\) ground-truth boxes \(G = \{g_1, \dots, g_N\}\) and \(K\) predicted boxes \(P = \{p_1, \dots, p_K\}\), we first compute an optimal one-to-one matching (e.g., via Hungarian matching). We then reward predictions that match successfully and have an Intersection-over-Union (IoU) greater than 0.5.

For each matched pair \((g, p)\) where \(g \in G, p \in P\), the reward is defined as: $
R_{\text{IoU}}(g,p) \;=\; \mathbf{1}_{\{\mathrm{IoU}(g,p) > 0.5\}}
$

(4) Continuous Score Reward.
This reward is applied to rules requiring continuous subjective scoring, specifically Aesthetic Attribute and Advertising Attribute.

Instead of using a binary 0/1 reward, we adopt a Gaussian-shaped function to softly reward predictions that closely match human ratings. This formulation encourages the model to produce fine-grained predictions that are as close as possible to the human-annotated ground-truth score \(\hat{s}\), rather than merely hitting a coarse interval.

Formally, let \(s\) be the model's predicted score and \(\hat{s}\) the human-provided reference score. The reward is defined as:

\begin{equation}
R_{\text{score}} = \exp\left( -\frac{(s - \hat{s})^2}{2\sigma^2} \right),
\end{equation}

Here, \(\sigma\) is a tunable standard deviation controlling the sharpness of the reward curve.

\textbf{Total Reward and Implementation Notes.}
Let \(\mathcal{A}\) be the set of active rewards for the current sample , with weights \(\alpha_i \geq 0\). To avoid scale drift due to different active subsets, we compute the total reward as a normalized weighted sum:
\begin{equation}
R_{\text{total}} = \frac{\sum_{i \in \mathcal{A}} \alpha_i R_i}{\sum_{i \in \mathcal{A}} \alpha_i}, 
\end{equation}

We always compute the general terms \(R_{\text{format}}\) and \(R_{\text{nonrep}}\), and then include rule-specific terms \(R_{\text{acc}}, R_{\text{IoU}}, R_{\text{tool}}, R_{\text{score}}\) only when applicable.

This reward framework allows the model to balance structural correctness, aesthetic accuracy, tool-grounded reasoning, and fine-grained subjective alignment. By integrating external tools into the CoT process and shaping outputs across multiple fidelity levels, the training process aligns the MLLM with the structure of aesthetic judgment.
 \section{Experiments}

\begin{table*}[t]
\centering

\setlength{\tabcolsep}{3pt}
\renewcommand{\arraystretch}{1.15}
\caption{\textbf{Performance across models under A$^3$-Law.}
The table reports results for the three stages \emph{Perceptual Attention}, \emph{Formal Interest}, and \emph{Desire Impact} with their subcriteria. 
For classification rules we show \textit{Acc}. 
For icon grounding we show \textit{mAP@0.5}. 
For continuous subjective prediction on \emph{Aesthetic Attribute} and \emph{Advertising Attribute} we show \textit{SRCC} and \textit{PLCC}. 
Models are grouped by type, including open source, open source with thinking, closed source, and closed source with thinking, with A$^3$-Align listed in the final row.}
\label{tab:a3law_models_resorted}
\resizebox{\textwidth}{!}{%
\begin{tabular}{l|ccc|ccc|c|cc|cc|cc}
\toprule
\textbf{Model} &
\multicolumn{3}{c|}{\textbf{\makecell{Perceptual\\Attention}}} &
\multicolumn{3}{c|}{\textbf{\makecell{Formal\\Interest}}} &
\multicolumn{7}{c}{\textbf{\makecell{Desire\\Impact}}} \\
\cmidrule(lr){2-4}\cmidrule(lr){5-7}\cmidrule(lr){8-14}
& \makecell{\small Image\\\small Fidelity}
& \makecell{\small Integration\\\small Realism}
& \makecell{\small Professional\\\small Polish}
& \makecell{\small Hue\\\small Adaptability}
& \makecell{\small Color\\\small Harmonization}
& \makecell{\small Layout\\\small Adaptability}
& \makecell{\small Copywriting\\\small Tone}
& \multicolumn{2}{c|}{\makecell{\small Promotional\\\small Iconography}}
& \multicolumn{2}{c|}{\makecell{\small Aesthetic\\\small Attribute}}
& \multicolumn{2}{c}{\makecell{\small Advertising\\\small Attribute}} \\
\cmidrule(lr){2-7}\cmidrule(lr){8-8}\cmidrule(lr){9-10}\cmidrule(lr){11-12}\cmidrule(lr){13-14}
& Acc$\uparrow$ & Acc$\uparrow$ & Acc$\uparrow$
& Acc$\uparrow$ & Acc$\uparrow$ & Acc$\uparrow$
& Acc$\uparrow$
& Acc$\uparrow$ & \multicolumn{1}{c|}{mAP@0.5$\uparrow$}
& SRCC$\uparrow$ & PLCC$\uparrow$
& SRCC$\uparrow$ & PLCC$\uparrow$ \\
\midrule
\multicolumn{14}{l}{\textit{Open-source Models}} \\
\midrule
Qwen3-VL-8B-Instruct \cite{qwen3technicalreport} & 0.454 & 0.491 & 0.463 & 0.491 & 0.444 & 0.472 & 0.611 & 0.157 & 0.011 & 0.564 & 0.562 & 0.533 & 0.528 \\
Gemma-3-27B-it \cite{team2025gemma} & 0.648 & 0.574 & 0.722 & 0.639 & 0.583 & 0.694 & 0.667 & 0.333 & 0.001 & 0.677 & 0.657 & 0.660 & 0.660 \\
Qwen3-VL-32B-Instruct \cite{qwen3technicalreport} & 0.602 & 0.583 & 0.518 & 0.694 & 0.574 & 0.482 & 0.880 & 0.259 & 0.172 & 0.640 & 0.617 & 0.728 & 0.701 \\
Qwen2.5-VL-72B-Instruct \cite{Qwen2.5-VL} & 0.509 & 0.380 & 0.481 & 0.676 & 0.463 & 0.518 & 0.824 & 0.343 & 0.032 & 0.544 & 0.512 & 0.628 & 0.597 \\
Glm-4.5v-106B \cite{vteam2025glm45vglm41vthinkingversatilemultimodal} & 0.500 & 0.509 & 0.486 & 0.615 & 0.523 & 0.509 & 0.650 & 0.107 & 0.069 & 0.652 & 0.618 & 0.661 & 0.650 \\
Llama-4-Scout-109B & 0.546 & 0.546 & 0.491 & 0.546 & 0.472 & 0.519 & 0.694 & 0.417 & 0.019 & 0.636 & 0.613 & 0.565 & 0.562 \\
Llama-4-Maverick-400B & 0.518 & 0.491 & 0.509 & 0.630 & 0.472 & 0.491 & 0.676 & 0.259 & 0.066 & 0.490 & 0.478 & 0.539 & 0.519 \\
\midrule
\multicolumn{14}{l}{\textit{Open-source Models - Thinking}} \\
\midrule
Qwen3-VL-32B-thinking & 0.556 & 0.481 & 0.481 & 0.676 & 0.500 & 0.481 & 0.870 & 0.222 & 0.110 & 0.607 & 0.599 & 0.642 & 0.618 \\
Qwen3-VL-235B-A22B-thinking & 0.546 & 0.602 & 0.630 & 0.741 & 0.630 & 0.639 & 0.769 & 0.120 & 0.141 & 0.667 & 0.613 & 0.669 & 0.651 \\
\midrule
\multicolumn{14}{l}{\textit{Closed-source Models}} \\
\midrule
ChatGPT 4.1 & 0.343 & 0.296 & 0.333 & 0.620 & 0.417 & 0.481 & 0.694 & 0.278 & 0.008 & 0.676 & 0.640 & 0.703 & 0.645 \\
ChatGPT 4o \cite{hurst2024gpt}& 0.518 & 0.529 & 0.472 & 0.620 & 0.481 & 0.537 & 0.778 & 0.222 & 0.001 & 0.680 & 0.648 & 0.781 & 0.756 \\
Mistral medium 3.1 \cite{mistral-medium-3-announce-2025} & 0.537 & 0.574 & 0.500 & 0.593 & 0.444 & 0.537 & 0.870 & 0.361 & 0.000 & 0.686 & 0.660 & 0.604 & 0.587 \\
Doubao 1.5 vision pro \cite{guo2025seed1} & 0.509 & 0.639 & 0.495 & 0.732 & 0.602 & 0.611 & 0.806 & 0.046 & 0.227 & 0.699 & 0.663 & 0.587 & 0.539 \\
Doubao seed 1.6 vision \cite{seed16vision-2025} & 0.481 & 0.528 & 0.467 & 0.546 & 0.472 & 0.472 & 0.704 & 0.417 & 0.032 & 0.680 & 0.633 & 0.527 & 0.458 \\
Grok 4 \cite{grok4-model-card-2025} & 0.491 & 0.611 & 0.602 & 0.722 & 0.426 & 0.528 & 0.768 & 0.454 & 0.002 & 0.698 & 0.674 & 0.690 & 0.672 \\
ChatGPT 5 \cite{gpt5-system-card-2025}& 0.509 & 0.648 & 0.500 & 0.694 & 0.500 & 0.537 & 0.917 & 0.250 & 0.013 & 0.732 & 0.698 & 0.752 & 0.703 \\
Claude Haiku 4.5 \cite{claude-haiku-45-2025}& 0.556 & 0.639 & 0.589 & 0.778 & 0.509 & 0.537 & 0.843 & 0.407 & 0.001 & 0.571 & 0.542 & 0.680 & 0.658 \\
Claude Sonnet 4.5 \cite{claude-sonnet-45-system-card-2025}& 0.546 & 0.528 & 0.608 & 0.732 & 0.611 & 0.561 & 0.824 & 0.472 & 0.030 & 0.725 & 0.691 & 0.689 & 0.677 \\
Claude Opus 4.1 \cite{claude-opus-41-2025} & 0.528 & 0.565 & 0.757 & 0.768 & 0.704 & 0.630 & 0.833 & 0.352 & 0.036 & 0.772 & 0.749 & 0.749 & 0.751 \\
Gemini 2.5 flash \cite{comanici2025gemini} & 0.667 & 0.531 & 0.708 & 0.633 & 0.473 & 0.568 & 0.680 & 0.140 & 0.013 & 0.704 & 0.676 & 0.626 & 0.622 \\
Gemini 2.5 pro & 0.750 & 0.732 & 0.806 & 0.806 & 0.509 & 0.750 & 0.870 & 0.463 & 0.027 & 0.743 & 0.744 & 0.704 & 0.686 \\
\midrule
\multicolumn{14}{l}{\textit{Closed-source Models - Thinking}} \\
\midrule
Doubao 1.5 vision pro-thinking & 0.472 & 0.482 & 0.467 & 0.676 & 0.444 & 0.472 & 0.833 & 0.102 & 0.151 & 0.724 & 0.673 & 0.716 & 0.660 \\
Doubao seed 1.6-thinking & 0.537 & 0.454 & 0.579 & 0.694 & 0.509 & 0.602 & 0.880 & 0.361 & 0.097 & 0.688 & 0.651 & 0.729 & 0.704 \\
ChatGPT o3 & 0.500 & 0.750 & 0.574 & 0.750 & 0.556 & 0.617 & 0.888 & 0.449 & 0.034 & 0.736 & 0.699 & 0.763 & 0.715 \\
ChatGPT o4 mini high & 0.537 & 0.750 & 0.648 & 0.822 & 0.574 & 0.608 & 0.879 & 0.780 & 0.036 & 0.711 & 0.667 & 0.720 & 0.702 \\
ChatGPT 5 high & 0.509 & 0.620 & 0.509 & 0.694 & 0.500 & 0.576 & 0.915 & 0.288 & 0.283 & 0.740 & 0.703 & 0.764 & 0.714 \\
Claude Haiku 4.5-thinking & 0.565 & 0.620 & 0.626 & 0.732 & 0.482 & 0.546 & 0.889 & 0.398 & 0.001 & 0.568 & 0.555 & 0.729 & 0.705 \\
Claude Sonnet 4.5-thinking & 0.556 & 0.546 & 0.689 & 0.757 & 0.574 & 0.626 & 0.815 & 0.472 & 0.024 & 0.732 & 0.697 & 0.713 & 0.696 \\
Claude Opus 4.1-thinking & 0.518 & 0.593 & 0.636 & 0.815 & 0.750 & 0.636 & 0.796 & 0.435 & 0.037 & 0.722 & 0.701 & 0.754 & 0.736 \\
Gemini 2.5 flash-thinking & 0.692 & 0.570 & 0.694 & 0.736 & 0.443 & 0.561 & 0.802 & 0.160 & 0.000 & 0.725 & 0.702 & 0.624 & 0.629 \\
Gemini 2.5 pro-thinking & 0.778 & 0.657 & 0.759 & 0.806 & 0.500 & 0.685 & 0.898 & 0.472 & 0.019 & 0.759 & 0.762 & 0.764 & 0.743 \\
\midrule
\textbf{A$^3$-Align} (Ours) &
\textbf{0.870} & \textbf{0.824} & \textbf{0.917} &
\textbf{0.824} & \textbf{0.880} & \textbf{0.870} &
\textbf{0.991} &
\textbf{0.926} & \textbf{0.701} &
\textbf{0.880} & \textbf{0.880} &
\textbf{0.838} & \textbf{0.836} \\
\bottomrule
\end{tabular}%
}
\end{table*}

\begin{table*}[t]
\centering

\setlength{\tabcolsep}{3pt}
\renewcommand{\arraystretch}{1.15}
\caption{Ablation analysis under A$^3$-Law. Accuracy is reported for classification rules, mAP@0.5 for icon grounding, and SRCC/PLCC for continuous subjective prediction.}
\label{tab:a3law_ablation}
\resizebox{\textwidth}{!}{%
\begin{tabular}{l|ccc|ccc|c|cc|cc|cc}
\toprule
\textbf{Method} &
  \multicolumn{3}{c|}{\textbf{\makecell{Perceptual\\Attention}}} &
  \multicolumn{3}{c|}{\textbf{\makecell{Formal\\Interest}}} &
  \multicolumn{7}{c}{\textbf{\makecell{Desire\\Impact}}} \\
\cmidrule(lr){2-4}\cmidrule(lr){5-7}\cmidrule(lr){8-14}
& \makecell{\small Image\\\small Fidelity}
& \makecell{\small Integration\\\small Realism}
& \makecell{\small Professional\\\small Polish}
& \makecell{\small Hue\\\small Adaptability}
& \makecell{\small Color\\\small Harmonization}
& \makecell{\small Layout\\\small Adaptability}
& \makecell{\small Copywriting\\\small Tone}
& \multicolumn{2}{c|}{\makecell{\small Promotional\\\small Iconography}}
& \multicolumn{2}{c|}{\makecell{\small Aesthetic\\\small Attribute}}
& \multicolumn{2}{c}{\makecell{\small Advertising\\\small Attribute}} \\
\cmidrule(lr){2-7}\cmidrule(lr){8-8}\cmidrule(lr){9-10}\cmidrule(lr){11-12}\cmidrule(lr){13-14}
& Acc$\uparrow$ & Acc$\uparrow$ & Acc$\uparrow$
& Acc$\uparrow$ & Acc$\uparrow$ & Acc$\uparrow$
& Acc$\uparrow$
& Acc$\uparrow$ & \multicolumn{1}{c|}{mAP@0.5$\uparrow$}
& SRCC$\uparrow$ & PLCC$\uparrow$
& SRCC$\uparrow$ & PLCC$\uparrow$ \\
\midrule
w/o CoT &
0.848 & 0.796 & 0.880 &
0.769 & 0.833 & 0.806 &
0.962 &
0.889 & 0.678 &
0.805 & 0.811 &
0.792 & 0.788 \\

w/o Accuracy Reward &
0.833 & 0.788 & 0.870 &
0.778 & 0.824 & 0.787 &
0.935 &
0.833 & 0.692 &
0.862 & 0.859 &
0.824 & 0.817 \\

w/o Tool Reward &
0.861 & 0.824 & 0.898 &
0.751 & 0.824 & 0.833 &
0.962 &
0.917 & 0.681 &
0.850 & 0.843 &
0.827 & 0.824 \\

w/o IoU Reward &
0.861 & 0.815 & 0.907 &
0.806 & 0.870 & 0.870 &
0.981 &
0.861 & 0.624 &
0.870 & 0.873 &
0.822 & 0.825 \\

w/o Continuous Reward &
0.843 & 0.806 & 0.898 &
0.806 & 0.852 & 0.870 &
0.981 &
0.917 & 0.690 &
0.820 & 0.815 &
0.804 & 0.803 \\

w/o Format Reward &
0.852 & 0.806 & 0.889 &
0.796 & 0.852 & 0.824 &
0.972 &
0.898 & 0.683 &
0.870 & 0.855 &
0.817 & 0.827 \\

w/o Non-Repeat Reward &
0.843 & 0.815 & 0.889 &
0.806 & 0.861 & 0.861 &
0.972 &
0.907 & 0.688 &
0.866 & 0.871 &
0.824 & 0.822 \\

\textbf{Ours (Full)} &
\textbf{0.870} & \textbf{0.824} & \textbf{0.917} &
\textbf{0.824} & \textbf{0.880} & \textbf{0.870} &
\textbf{0.991} &
\textbf{0.926} & \textbf{0.701} &
\textbf{0.880} & \textbf{0.880} &
\textbf{0.838} & \textbf{0.836} \\
\bottomrule
\end{tabular}%
}
\end{table*}

\subsection{Experimental Setup}

\textbf{Training \& Evaluation} We select Qwen3-VL-8B-Instruct \cite{qwen3technicalreport} as the base multimodal model, with all experiments conducted on an 8$\times$H200 GPU. Our training process encompasses two core stages: first, we perform Supervised Fine-Tuning (SFT), followed by reinforcement learning using Group Relative Policy Optimization (GRPO) integrated with a multi-signal feedback mechanism. During the GRPO stage, key parameters are set to $n=4$ and $\sigma=1.237$. For data partitioning, we split the annotated dataset into training, internal validation, and held-out test sets using an 8:1:1 ratio, ensuring no overlap between the subsets.All model performance evaluations are conducted on a unified test set. We employ specific metrics for different evaluation dimensions: for objective rules, we report Accuracy; for Promotional Iconography, we report both Accuracy and mAP@0.5; and for the two subjective rule categories, we report the Spearman's Rank Correlation Coefficient (SRCC) and the Pearson's Linear Correlation Coefficient (PLCC).

\subsection{Results}

\textbf{A$^3$-Bench. }The proposed A$^3$-Bench covers over 20 mainstream model variants, including both state-of-the-art (SOTA) open-source and closed-source models, and innovatively compares their performance under ``non-thinking" (without reasoning chain) and ``thinking" inference modes.

Overall, closed-source models exhibit consistently stronger performance compared to open-source counterparts. For instance, in the three perceptual attention subtasks, the best-performing closed-source model, Gemini-2.5-pro \cite{comanici2025gemini}, achieves scores of 0.750, 0.732, and 0.806, significantly surpassing the best open-source model, Gemma-3-27B \cite{team2025gemma} (0.648, 0.574, and 0.722 respectively). Similarly, in subjective aesthetic evaluation tasks, the top closed-source model (Claude Opus 4.1 \cite{claude-opus-41-2025}, 0.772) clearly outperforms the top open-source model (Gemma-3-27B, 0.677). Notably, both open-source and closed-source models struggle with promotional icon localization. The best open-source mAP is only 0.172, while the highest closed-source mAP is 0.283. This indicates a limitation among current models: they can determine whether a promotional icon exists but struggle with accurate spatial localization.

Analysis of models adopting a ``thinking'' approach reveals differing effects across model types. For open-source models, enabling ``thinking'' tends to be unstable or even harmful; for example, Qwen3-VL-32B-Instruct shows declines in subjective rating correlations and a slight reduction in icon localization mAP. Conversely, closed-source models consistently benefit from thinking-mode inference. For instance, Gemini-2.5-pro’s subjective rating correlations improve from 0.743 to 0.759 after enabling it. Nonetheless, even these moderate improvements do not address the weak localization performance, indicating that although the ``thinking'' mechanism facilitates better evidence integration, it remains insufficient for solving localization tasks.

Our proposed A$^3$-Align significantly outperforms all existing models across every evaluation metric. Specifically, on perceptual attention subtasks, A$^3$-Align achieves scores of 0.870, 0.824, and 0.917, improving upon the strongest baselines by margins of 0.092, 0.074, and 0.111, respectively. In formal interest subtasks, A$^3$-Align's advantage is even more pronounced, with improvements of 0.176 and 0.120 on color harmonization and layout adaptability compared to the strongest existing models. Crucially, on the challenging promotional icon localization task, A$^3$-Align not only accurately classifies the presence of icons (accuracy improved to 0.926), but also significantly boosts spatial precision, achieving an mAP of 0.701—more than double the best closed-source baseline (0.283). Additionally, in continuous subjective scoring tasks, A$^3$-Align attains correlation scores of 0.880, marking an improvement greater than 0.108 over prior best-performing models.

Overall, these substantial performance gains can be attributed to our specialized training designs: the model explicitly leverages tool utilization rewards for evidence-based reasoning, employs IoU-based rewards to fundamentally enhance spatial localization, and uses continuous score rewards to align precisely with human aesthetic judgments. Consequently, A$^3$-Align not only delivers SOTA performance across individual tasks but also internalizes the core aesthetic principles defined by the A$^3$-Law framework.

\begin{figure}[t]
  \centering
   \includegraphics[width=1.05\linewidth]{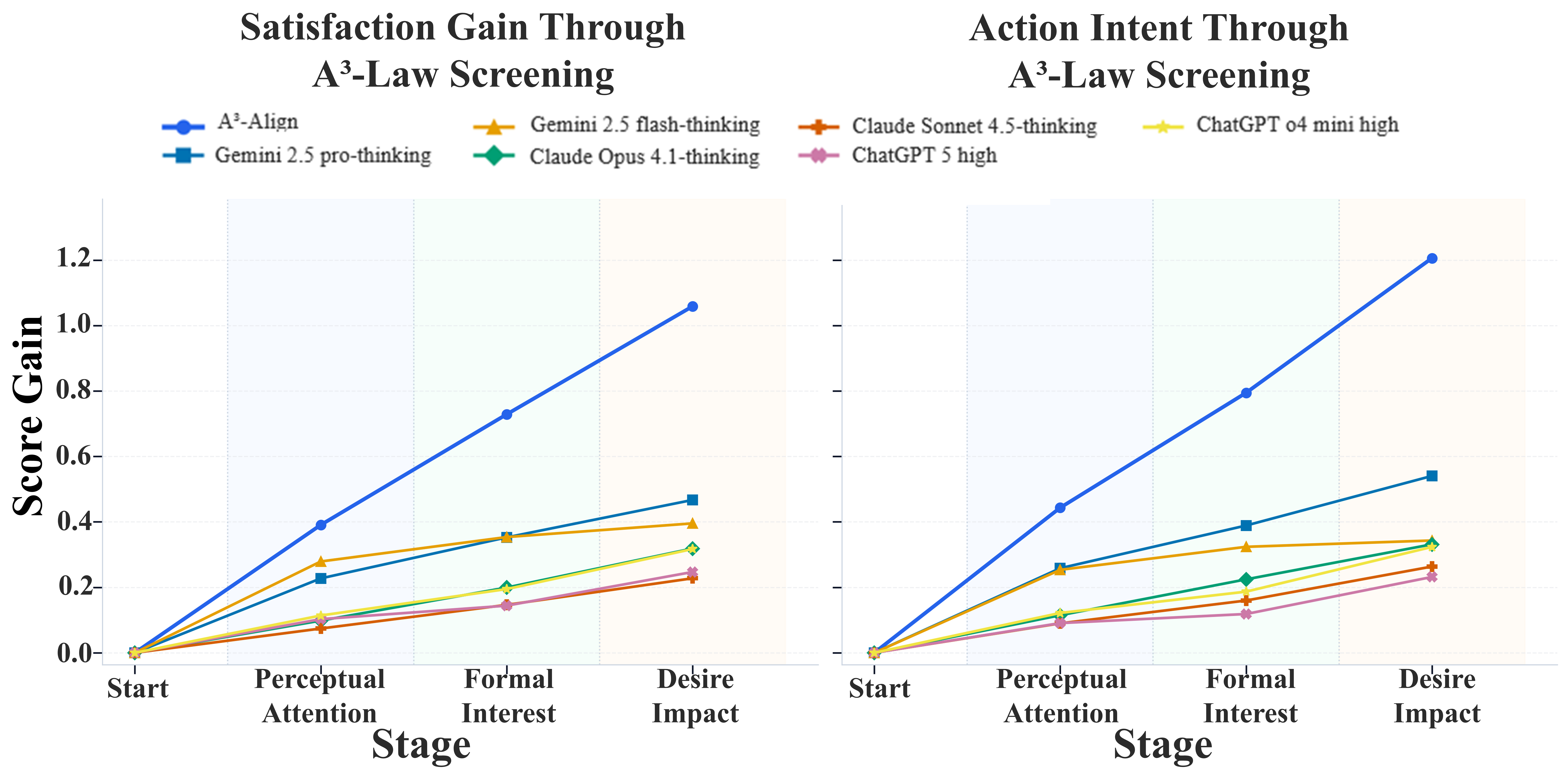}

   \caption{\textbf{Satisfaction and Action Intent Gains Through A$^3$-Law Screening.} 
The figure shows the cumulative score gains in \emph{Satisfaction} (left) and \emph{Action Intent} (right) across the three stages of A$^3$-Law screening: \emph{Perceptual Attention}, \emph{Formal Interest}, and \emph{Desire Impact}. }
   \label{fig:figure-4}
\end{figure}
\textbf{Application I: Quality Advertisement Selection}
To validate the effectiveness of A$^3$-Law in selecting high-quality advertisements, five raters scored 100 images from the A$^3$-Dataset on 1 to 5 scales for \textbf{Satisfaction} and \textbf{Action Intent}. Figure \ref{fig:figure-4} reports cumulative gains from the unfiltered baseline \emph{Start} (S = 2.956, A = 2.994) through \emph{Perceptual Attention}, \emph{Formal Interest}, and \emph{Desire Impact}. Gains are monotonic for all models, confirming the value of A$^3$-Law. A$^3$-Align yields the largest improvement, reaching \(+1.05\) in Satisfaction \((35.5\%)\) and \(+1.20\) in Action Intent \((40.1\%)\) at \emph{Desire Impact}, which is \(3\times\) the mean gain of other models. The lift in Action Intent exceeds the lift in Satisfaction, indicating that A$^3$-Law improves perceived quality and more strongly increases the propensity to act.

\begin{figure}[t]
  \centering
   \includegraphics[width=0.99\linewidth]{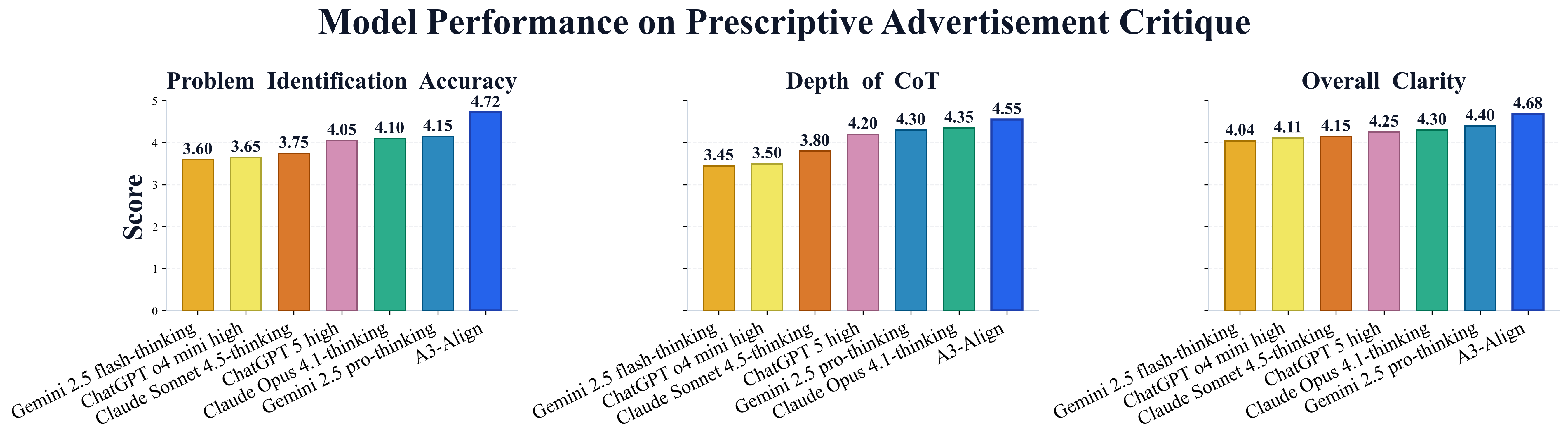}
   \caption{\textbf{Evaluation of Problem Identification Accuracy, Depth of CoT, and Overall Clarity.} 
The figure presents the scores of various models on three evaluation dimensions: \emph{Problem Identification Accuracy}, \emph{Depth of Chain of Thought (CoT)}, and \emph{Overall Clarity}. }
   \label{fig:figure-5}
\end{figure}
\textbf{Application II: Prescriptive Advertisement Critique}
Figure \ref{fig:figure-5} compares three open-ended critique dimensions—problem identification accuracy, chain-of-thought depth, and overall clarity—averaged over 30 images. A$^3$-Align leads on all axes (mean 4.65), surpassing the next best by +0.37 (+8.6\%). The largest margin appears in problem identification (+0.57), while depth (+0.20) and clarity (+0.28) remain consistently higher, indicating that A$^3$-Align not only detects the right issues but also articulates deeper and clearer prescriptive guidance. Further evaluations demonstrating $A^3$-Align's robust generalization on the external AdImageNet \cite{brendan_adimagenet} dataset and its potential for generative AI steering are detailed in the supplementary material.

\textbf{Limitations and Failure Analysis.} Despite its strong performance, $A^3$-Align exhibits two primary failure modes: \textbf{(1)} Spatial crowding, where dense layouts trigger icon hallucinations or merging; and \textbf{(2)} Attribute miscalibration, where the model occasionally underestimates minimalist premium designs by over-associating visual simplicity with weaker advertising effectiveness.

\subsection{Ablation Study}

Ablation studies confirm the roles of explicit CoT and targeted RL rewards. Without CoT, the model shows limited drops on basic perceptual tasks but much larger declines on subjective alignment tasks, with Aesthetic Attribute SRCC decreasing from 0.880 to 0.805. The RL rewards are also complementary: removing the Accuracy, IoU, and Tool rewards lowers Promotional Iconography accuracy by 0.093, grounding mAP@0.5 by 0.077, and Hue Adaptability accuracy by 0.073, respectively. The Continuous Reward is crucial for subjective prediction, while the Format and Non-Repeat rewards provide smaller gains. Together, these components produce the best overall performance. Furthermore, we observe a hierarchical learning process in which format and non-repeat rewards converge first, followed by tool usage, accuracy, IoU, and finally the continuous score reward.

\section{Conclusion}
In this paper, we introduce A$^3$ to address the subjectivity and limited scalability of advertising image evaluation by unifying a theory (A$^3$-Law), a dataset (A$^3$-Dataset), a model (A$^3$-Align), and a benchmark (A$^3$-Bench). 
Our A$^3$-Law defines three stages---%
\emph{Perceptual Attention}, \emph{Formal Interest}, and \emph{Desire Impact}---%
which guided the creation of A$^3$-Dataset (30K images, 120K instruction pairs with Chain of Thought). 
The A$^3$-Align model, trained with CoT-guided learning, demonstrates superior adherence to the law and transfers effectively to quality selection and prescriptive critique. 
This work shows broad potential for creative advertising aesthetic assessment.

\section{Acknowledgment}
This work was supported by the New Generation Artificial Intelligence National Science and Technology Major Project (Nos. 2025ZD0124104 and P25KK00221), in collaboration with the Shanghai Artificial Intelligence Laboratory, and by the National Natural Science Foundation of China (Grant Nos. 62571324, 62501337, and 62225112).

{
    \small
    \bibliographystyle{ieeenat_fullname}
    \bibliography{main}

@String(TOG= {ACM Trans. Graph.})

@String(ICME = {Int. Conf. Multimedia and Expo})

@String(IJCAI = {IJCAI})

@String(AAAI = {AAAI})

@String(TOG   = {ACM TOG})

@String(ICME  =	{ICME})

@article{ha2008integrated,
  title={An integrated model of advertising clutter in offline and online media},
  author={Ha, Louisa and McCann, Kim},
  journal={International Journal of Advertising},
  volume={27},
  number={4},
  pages={569--592},
  year={2008},
  publisher={Taylor \& Francis}
}

@article{rehman2023marketing,
  title={Marketing overload: The impact of information overload on brand recall (a case study of students of the University of Swat)},
  author={Rehman, Abdur and Ahmad, Ishfaq and Amin, Kashif and Noor, Nayab and Rehman, Asim},
  journal={Journal of Social Sciences Review},
  volume={3},
  number={2},
  pages={70--78},
  year={2023}
}

@incollection{ha2017digital,
  title={Digital advertising clutter in an age of mobile media},
  author={Ha, Louisa},
  booktitle={Digital advertising},
  pages={69--85},
  year={2017},
  publisher={Routledge}
}

@inproceedings{murray2012ava,
  title={AVA: A large-scale database for aesthetic visual analysis},
  author={Murray, Naila and Marchesotti, Luca and Perronnin, Florent},
  booktitle={2012 IEEE conference on computer vision and pattern recognition},
  pages={2408--2415},
  year={2012},
  organization={IEEE}
}

@article{li2024computer,
  title={Computer Vision Models for Image Analysis in Advertising Research},
  author={Li, Hairong and Zhang, Nan},
  journal={Journal of Advertising},
  volume={53},
  number={5},
  pages={771--790},
  year={2024},
  publisher={Taylor \& Francis}
}

@article{jang2022modeling,
  title={Modeling, Quantifying, and Predicting Subjectivity of Image Aesthetics},
  author={Jang, Hyeongnam and Lee, Yeejin and Lee, Jong-Seok},
  journal={arXiv preprint arXiv:2208.09666},
  year={2022}
}

@inproceedings{wang2019aspect,
  title={Aspect-ratio-preserving multi-patch image aesthetics score prediction},
  author={Wang, Lijie and Wang, Xueting and Yamasaki, Toshihiko and Aizawa, Kiyoharu},
  booktitle={Proceedings of the IEEE/CVF Conference on Computer Vision and Pattern Recognition Workshops},
  pages={0--0},
  year={2019}
}

@inproceedings{zhou2016predicting,
  title={Predicting pre-click quality for native advertisements},
  author={Zhou, Ke and Redi, Miriam and Haines, Andrew and Lalmas, Mounia},
  booktitle={Proceedings of the 25th International Conference on World Wide Web},
  pages={299--310},
  year={2016}
}

@inproceedings{dutt2024explainable,
  title={Explainable Digital Creatives Performance Monitoring using Deep Feature Attribution},
  author={Dutt, Varun and Hadjigeorgiou, Demetris and Galan, Lucas and Doctor, Faiyaz and Barakat, Lina and Isaacs, Kate},
  booktitle={2024 19th Annual System of Systems Engineering Conference (SoSE)},
  pages={134--139},
  year={2024},
  organization={IEEE}
}

@inproceedings{yue2024mmmu,
  title={Mmmu: A massive multi-discipline multimodal understanding and reasoning benchmark for expert agi},
  author={Yue, Xiang and Ni, Yuansheng and Zhang, Kai and Zheng, Tianyu and Liu, Ruoqi and Zhang, Ge and Stevens, Samuel and Jiang, Dongfu and Ren, Weiming and Sun, Yuxuan and others},
  booktitle={Proceedings of the IEEE/CVF Conference on Computer Vision and Pattern Recognition},
  pages={9556--9567},
  year={2024}
}

@inproceedings{liu2024mmbench,
  title={Mmbench: Is your multi-modal model an all-around player?},
  author={Liu, Yuan and Duan, Haodong and Zhang, Yuanhan and Li, Bo and Zhang, Songyang and Zhao, Wangbo and Yuan, Yike and Wang, Jiaqi and He, Conghui and Liu, Ziwei and others},
  booktitle={European conference on computer vision},
  pages={216--233},
  year={2024},
  organization={Springer}
}

@article{chen2025reasoning,
  title={Reasoning Models Don't Always Say What They Think},
  author={Chen, Yanda and Benton, Joe and Radhakrishnan, Ansh and Uesato, Jonathan and Denison, Carson and Schulman, John and Somani, Arushi and Hase, Peter and Wagner, Misha and Roger, Fabien and others},
  journal={arXiv preprint arXiv:2505.05410},
  year={2025}
}

@article{turpin2023language,
  title={Language models don't always say what they think: Unfaithful explanations in chain-of-thought prompting},
  author={Turpin, Miles and Michael, Julian and Perez, Ethan and Bowman, Samuel},
  journal={Advances in Neural Information Processing Systems},
  volume={36},
  pages={74952--74965},
  year={2023}
}

@article{baker2025monitoring,
  title={Monitoring reasoning models for misbehavior and the risks of promoting obfuscation},
  author={Baker, Bowen and Huizinga, Joost and Gao, Leo and Dou, Zehao and Guan, Melody Y and Madry, Aleksander and Zaremba, Wojciech and Pachocki, Jakub and Farhi, David},
  journal={arXiv preprint arXiv:2503.11926},
  year={2025}
}

@inproceedings{guan2024hallusionbench,
  title={Hallusionbench: an advanced diagnostic suite for entangled language hallucination and visual illusion in large vision-language models},
  author={Guan, Tianrui and Liu, Fuxiao and Wu, Xiyang and Xian, Ruiqi and Li, Zongxia and Liu, Xiaoyu and Wang, Xijun and Chen, Lichang and Huang, Furong and Yacoob, Yaser and others},
  booktitle={Proceedings of the IEEE/CVF Conference on Computer Vision and Pattern Recognition},
  pages={14375--14385},
  year={2024}
}

@article{strong1925theories,
  title={Theories of selling.},
  author={Strong Jr, Edward K},
  journal={Journal of applied psychology},
  volume={9},
  number={1},
  pages={75},
  year={1925},
  publisher={American Psychological Association}
}

@inproceedings{azimi2012impact,
  title={The impact of visual appearance on user response in online display advertising},
  author={Azimi, Javad and Zhang, Ruofei and Zhou, Yang and Navalpakkam, Vidhya and Mao, Jianchang and Fern, Xiaoli},
  booktitle={proceedings of the 21st international conference on World Wide Web},
  pages={457--458},
  year={2012}
}

@inproceedings{azimi2012visual,
  title={Visual appearance of display ads and its effect on click through rate},
  author={Azimi, Javad and Zhang, Ruofei and Zhou, Yang and Navalpakkam, Vidhya and Mao, Jianchang and Fern, Xiaoli},
  booktitle={Proceedings of the 21st ACM international conference on Information and knowledge management},
  pages={495--504},
  year={2012}
}

@article{bai2025comprehensive,
  title={A comprehensive survey on advertising click-through rate prediction algorithm},
  author={Bai, Jing and Geng, Xinyu and Deng, Jiaqi and Xia, Zhen and Jiang, Hongxia and Yan, Guoqiang and Liang, Jing},
  journal={The Knowledge Engineering Review},
  volume={40},
  pages={e3},
  year={2025},
  publisher={Cambridge University Press}
}

@article{talebi2018nima,
  title={NIMA: Neural image assessment},
  author={Talebi, Hossein and Milanfar, Peyman},
  journal={IEEE transactions on image processing},
  volume={27},
  number={8},
  pages={3998--4011},
  year={2018},
  publisher={IEEE}
}

@inproceedings{he2022rethinking,
  title={Rethinking Image Aesthetics Assessment: Models, Datasets and Benchmarks.},
  author={He, Shuai and Zhang, Yongchang and Xie, Rui and Jiang, Dongxiang and Ming, Anlong},
  booktitle={IJCAI},
  pages={942--948},
  year={2022}
}

@inproceedings{yi2023towards,
  title={Towards artistic image aesthetics assessment: a large-scale dataset and a new method},
  author={Yi, Ran and Tian, Haoyuan and Gu, Zhihao and Lai, Yu-Kun and Rosin, Paul L},
  booktitle={Proceedings of the IEEE/CVF Conference on Computer Vision and Pattern Recognition},
  pages={22388--22397},
  year={2023}
}

@inproceedings{kalra2020understanding,
  title={Understanding advertisements with BERT},
  author={Kalra, Kanika and Kurma, Bhargav and Sreelatha, Silpa Vadakkeeveetil and Patwardhan, Manasi and Karande, Shirish},
  booktitle={Proceedings of the 58th Annual Meeting of the Association for Computational Linguistics},
  pages={7542--7547},
  year={2020}
}

@article{dey2018don,
  title={Don't only Feel Read: Using Scene text to understand advertisements},
  author={Dey, Arka Ujjal and Ghosh, Suman K and Valveny, Ernest},
  journal={arXiv preprint arXiv:1806.08279},
  year={2018}
}

@inproceedings{savchenko2020ad,
  title={Ad lingua: Text classification improves symbolism prediction in image advertisements},
  author={Savchenko, Andrey and Alekseev, Anton and Kwon, Sejeong and Tutubalina, Elena and Myasnikov, Evgeny and Nikolenko, Sergey},
  booktitle={Proceedings of the 28th International Conference on Computational Linguistics},
  pages={1886--1892},
  year={2020}
}

@incollection{cohen2006color,
  title={Color harmonization},
  author={Cohen-Or, Daniel and Sorkine, Olga and Gal, Ran and Leyvand, Tommer and Xu, Ying-Qing},
  booktitle={ACM SIGGRAPH 2006 Papers},
  pages={624--630},
  year={2006}
}

@inproceedings{hasler2003measuring,
  title={Measuring colorfulness in natural images},
  author={Hasler, David and Suesstrunk, Sabine E},
  booktitle={Human vision and electronic imaging VIII},
  volume={5007},
  pages={87--95},
  year={2003},
  organization={SPIE}
}

@article{li2019layoutgan,
  title={Layoutgan: Generating graphic layouts with wireframe discriminators},
  author={Li, Jianan and Yang, Jimei and Hertzmann, Aaron and Zhang, Jianming and Xu, Tingfa},
  journal={arXiv preprint arXiv:1901.06767},
  year={2019}
}

@inproceedings{lee2020neural,
  title={Neural design network: Graphic layout generation with constraints},
  author={Lee, Hsin-Ying and Jiang, Lu and Essa, Irfan and Le, Phuong B and Gong, Haifeng and Yang, Ming-Hsuan and Yang, Weilong},
  booktitle={European conference on computer vision},
  pages={491--506},
  year={2020},
  organization={Springer}
}

@article{zheng2019content,
  title={Content-aware generative modeling of graphic design layouts},
  author={Zheng, Xinru and Qiao, Xiaotian and Cao, Ying and Lau, Rynson WH},
  journal={ACM Transactions on Graphics (TOG)},
  volume={38},
  number={4},
  pages={1--15},
  year={2019},
  publisher={ACM New York, NY, USA}
}

@inproceedings{schultze2023explaining,
  title={Explaining image aesthetics assessment: An interactive approach},
  author={Schultze, Sven and With{\"o}ft, Ani and Abdenebaoui, Larbi and Boll, Susanne},
  booktitle={Proceedings of the 2023 ACM International Conference on Multimedia Retrieval},
  pages={20--28},
  year={2023}
}

@article{tong2022interpretable,
  title={An interpretable approach for automatic aesthetic assessment of remote sensing images},
  author={Tong, Jingru and Zhang, Guo and Kong, Peijie and Rao, Yu and Wei, Zhengkai and Cui, Hao and Guan, Qing},
  journal={Frontiers in Computational Neuroscience},
  volume={16},
  pages={1077439},
  year={2022},
  publisher={Frontiers Media SA}
}

@article{santos2024towards,
  title={Towards Robust Evaluation of Aesthetic and Photographic Quality Metrics: Insights from a Comprehensive Dataset},
  author={Santos, Iria and Casal, Miguel A and Correia, Jo{\~a}o and Torrente-Pati{\~n}o, {\'A}lvaro and Machado, Penousal and Romero, Juan},
  journal={Complexity},
  volume={2024},
  number={1},
  pages={8223586},
  year={2024},
  publisher={Wiley Online Library}
}

@inproceedings{hussain2017automatic,
  title={Automatic understanding of image and video advertisements},
  author={Hussain, Zaeem and Zhang, Mingda and Zhang, Xiaozhong and Ye, Keren and Thomas, Christopher and Agha, Zuha and Ong, Nathan and Kovashka, Adriana},
  booktitle={Proceedings of the IEEE conference on computer vision and pattern recognition},
  pages={1705--1715},
  year={2017}
}

@article{ahuja2018understanding,
  title={Understanding visual ads by aligning symbols and objects using co-attention},
  author={Ahuja, Karuna and Sikka, Karan and Roy, Anirban and Divakaran, Ajay},
  journal={arXiv preprint arXiv:1807.01448},
  year={2018}
}

@article{hewitt2024instruction,
  title={Instruction following without instruction tuning},
  author={Hewitt, John and Liu, Nelson F and Liang, Percy and Manning, Christopher D},
  journal={arXiv preprint arXiv:2409.14254},
  year={2024}
}

@inproceedings{chen2024internvl,
  title={Internvl: Scaling up vision foundation models and aligning for generic visual-linguistic tasks},
  author={Chen, Zhe and Wu, Jiannan and Wang, Wenhai and Su, Weijie and Chen, Guo and Xing, Sen and Zhong, Muyan and Zhang, Qinglong and Zhu, Xizhou and Lu, Lewei and others},
  booktitle={Proceedings of the IEEE/CVF conference on computer vision and pattern recognition},
  pages={24185--24198},
  year={2024}
}

@inproceedings{kong2016photo,
  title={Photo aesthetics ranking network with attributes and content adaptation},
  author={Kong, Shu and Shen, Xiaohui and Lin, Zhe and Mech, Radomir and Fowlkes, Charless},
  booktitle={European conference on computer vision},
  pages={662--679},
  year={2016},
  organization={Springer}
}

@article{huang2024aesbench,
  title={Aesbench: An expert benchmark for multimodal large language models on image aesthetics perception},
  author={Huang, Yipo and Yuan, Quan and Sheng, Xiangfei and Yang, Zhichao and Wu, Haoning and Chen, Pengfei and Yang, Yuzhe and Li, Leida and Lin, Weisi},
  journal={arXiv preprint arXiv:2401.08276},
  year={2024}
}

@inproceedings{qi2025photographer,
  title={The Photographer's Eye: Teaching Multimodal Large Language Models to See, and Critique Like Photographers},
  author={Qi, Daiqing and Zhao, Handong and Shi, Jing and Jenni, Simon and Fan, Yifei and Dernoncourt, Franck and Cohen, Scott and Li, Sheng},
  booktitle={Proceedings of the Computer Vision and Pattern Recognition Conference},
  pages={24807--24816},
  year={2025}
}

@inproceedings{jia2023kafa,
  title={KAFA: Rethinking image ad understanding with knowledge-augmented feature adaptation of vision-language models},
  author={Jia, Zhiwei and Narayana, Pradyumna and Akula, Arjun and Pruthi, Garima and Su, Hao and Basu, Sugato and Jampani, Varun},
  booktitle={Proceedings of the 61st Annual Meeting of the Association for Computational Linguistics (Volume 5: Industry Track)},
  pages={772--785},
  year={2023}
}

@article{barry1987development,
  title={The development of the hierarchy of effects: An historical perspective},
  author={Barry, Thomas E},
  journal={Current issues and Research in Advertising},
  volume={10},
  number={1-2},
  pages={251--295},
  year={1987},
  publisher={Taylor \& Francis}
}

@book{green1966signal,
  title={Signal detection theory and psychophysics},
  author={Green, David Marvin and Swets, John A and others},
  volume={1},
  year={1966},
  publisher={Wiley New York}
}

@book{hautus2021detection,
  title={Detection theory: A user's guide},
  author={Hautus, Michael J and Macmillan, Neil A and Creelman, C Douglas},
  year={2021},
  publisher={Routledge}
}

@article{shannon1948mathematical,
  title={A mathematical theory of communication},
  author={Shannon, Claude E},
  journal={The Bell system technical journal},
  volume={27},
  number={3},
  pages={379--423},
  year={1948},
  publisher={Nokia Bell Labs}
}

@article{potter2014detecting,
  title={Detecting meaning in RSVP at 13 ms per picture},
  author={Potter, Mary C and Wyble, Brad and Hagmann, Carl Erick and McCourt, Emily S},
  journal={Attention, Perception, \& Psychophysics},
  volume={76},
  number={2},
  pages={270--279},
  year={2014},
  publisher={Springer}
}

@article{thorpe1996speed,
  title={Speed of processing in the human visual system},
  author={Thorpe, Simon and Fize, Denis and Marlot, Catherine},
  journal={nature},
  volume={381},
  number={6582},
  pages={520--522},
  year={1996},
  publisher={Nature Publishing Group UK London}
}

@article{johnston1985perceptual,
  title={Perceptual fluency and recognition judgments.},
  author={Johnston, William A and Dark, Veronica J and Jacoby, Larry L},
  journal={Journal of Experimental Psychology: Learning, Memory, and Cognition},
  volume={11},
  number={1},
  pages={3},
  year={1985},
  publisher={American Psychological Association}
}

@article{reber1999effects,
  title={Effects of perceptual fluency on judgments of truth},
  author={Reber, Rolf and Schwarz, Norbert},
  journal={Consciousness and cognition},
  volume={8},
  number={3},
  pages={338--342},
  year={1999},
  publisher={Elsevier}
}

@article{silvia2005interesting,
  title={What is interesting? Exploring the appraisal structure of interest.},
  author={Silvia, Paul J},
  journal={Emotion},
  volume={5},
  number={1},
  pages={89},
  year={2005},
  publisher={American Psychological Association}
}

@article{berlyne1970novelty,
  title={Novelty, complexity, and hedonic value},
  author={Berlyne, Daniel E},
  journal={Perception \& psychophysics},
  volume={8},
  number={5},
  pages={279--286},
  year={1970},
  publisher={Springer}
}

@article{treisman1980feature,
  title={A feature-integration theory of attention},
  author={Treisman, Anne M and Gelade, Garry},
  journal={Cognitive psychology},
  volume={12},
  number={1},
  pages={97--136},
  year={1980},
  publisher={Elsevier}
}

@article{wolfe2017five,
  title={Five factors that guide attention in visual search},
  author={Wolfe, Jeremy M and Horowitz, Todd S},
  journal={Nature human behaviour},
  volume={1},
  number={3},
  pages={0058},
  year={2017},
  publisher={Nature Publishing Group UK London}
}

@article{ou2006colour,
  title={A colour harmony model for two-colour combinations},
  author={Ou, Li-Chen and Luo, M Ronnier},
  journal={Color Research \& Application: Endorsed by Inter-Society Color Council, The Colour Group (Great Britain), Canadian Society for Color, Color Science Association of Japan, Dutch Society for the Study of Color, The Swedish Colour Centre Foundation, Colour Society of Australia, Centre Fran{\c{c}}ais de la Couleur},
  volume={31},
  number={3},
  pages={191--204},
  year={2006},
  publisher={Wiley Online Library}
}

@book{koffka2013principles,
  title={Principles of Gestalt psychology},
  author={Koffka, Kurt},
  year={2013},
  publisher={routledge}
}

@article{tufte1991envisioning,
  title={Envisioning information},
  author={Tufte, Edward R},
  journal={Optometry and Vision Science},
  volume={68},
  number={4},
  pages={322--324},
  year={1991},
  publisher={LWW}
}

@article{barthes1985rhetoric,
  title={Rhetoric of the Image},
  author={Barthes, Roland},
  journal={Semiotics: An introductory anthology},
  pages={192--205},
  year={1985},
  publisher={Bloomington: Indiana University Press}
}

@article{mick1986consumer,
  title={Consumer research and semiotics: Exploring the morphology of signs, symbols, and significance},
  author={Mick, David Glen},
  journal={Journal of consumer research},
  volume={13},
  number={2},
  pages={196--213},
  year={1986},
  publisher={The University of Chicago Press}
}

@article{fishwick2004emotional,
  title={Emotional design: why we love (or hate) everyday things},
  author={Fishwick, Marshall},
  journal={The Journal of American Culture},
  volume={27},
  number={2},
  pages={234},
  year={2004},
  publisher={Blackwell Publishing Ltd.}
}

@book{scherer2001appraisal,
  title={Appraisal processes in emotion: Theory, methods, research},
  author={Scherer, Klaus R and Schorr, Angela and Johnstone, Tom},
  year={2001},
  publisher={Oxford University Press}
}

@article{norman2002emotion,
  title={Emotion \& design: attractive things work better},
  author={Norman, Don},
  journal={interactions},
  volume={9},
  number={4},
  pages={36--42},
  year={2002},
  publisher={ACM New York, NY, USA}
}

@article{wei2025deepseek,
  title={DeepSeek-OCR: Contexts Optical Compression},
  author={Wei, Haoran and Sun, Yaofeng and Li, Yukun},
  journal={arXiv preprint arXiv:2510.18234},
  year={2025}
}

@article{shao2024deepseekmath,
  title={Deepseekmath: Pushing the limits of mathematical reasoning in open language models},
  author={Shao, Zhihong and Wang, Peiyi and Zhu, Qihao and Xu, Runxin and Song, Junxiao and Bi, Xiao and Zhang, Haowei and Zhang, Mingchuan and Li, YK and Wu, Yang and others},
  journal={arXiv preprint arXiv:2402.03300},
  year={2024}
}

@article{comanici2025gemini,
  title={Gemini 2.5: Pushing the frontier with advanced reasoning, multimodality, long context, and next generation agentic capabilities},
  author={Comanici, Gheorghe and Bieber, Eric and Schaekermann, Mike and Pasupat, Ice and Sachdeva, Noveen and Dhillon, Inderjit and Blistein, Marcel and Ram, Ori and Zhang, Dan and Rosen, Evan and others},
  journal={arXiv preprint arXiv:2507.06261},
  year={2025}
}

@article{team2025gemma,
  title={Gemma 3 technical report},
  author={Team, Gemma and Kamath, Aishwarya and Ferret, Johan and Pathak, Shreya and Vieillard, Nino and Merhej, Ramona and Perrin, Sarah and Matejovicova, Tatiana and Ram{\'e}, Alexandre and Rivi{\`e}re, Morgane and others},
  journal={arXiv preprint arXiv:2503.19786},
  year={2025}
}

@misc{qwen3technicalreport,
      title={Qwen3 Technical Report}, 
      author={Qwen Team},
      year={2025},
      eprint={2505.09388},
      archivePrefix={arXiv},
      primaryClass={cs.CL},
      url={https://arxiv.org/abs/2505.09388}, 
}

@article{Qwen2.5-VL,
  title={Qwen2.5-VL Technical Report},
  author={Bai, Shuai and Chen, Keqin and Liu, Xuejing and Wang, Jialin and Ge, Wenbin and Song, Sibo and Dang, Kai and Wang, Peng and Wang, Shijie and Tang, Jun and Zhong, Humen and Zhu, Yuanzhi and Yang, Mingkun and Li, Zhaohai and Wan, Jianqiang and Wang, Pengfei and Ding, Wei and Fu, Zheren and Xu, Yiheng and Ye, Jiabo and Zhang, Xi and Xie, Tianbao and Cheng, Zesen and Zhang, Hang and Yang, Zhibo and Xu, Haiyang and Lin, Junyang},
  journal={arXiv preprint arXiv:2502.13923},
  year={2025}
}

@misc{vteam2025glm45vglm41vthinkingversatilemultimodal,
      title={GLM-4.5V and GLM-4.1V-Thinking: Towards Versatile Multimodal Reasoning with Scalable Reinforcement Learning}, 
      author={V Team and Wenyi Hong and Wenmeng Yu and Xiaotao Gu and Guo Wang and Guobing Gan and Haomiao Tang and Jiale Cheng and Ji Qi and Junhui Ji and Lihang Pan and Shuaiqi Duan and Weihan Wang and Yan Wang and Yean Cheng and Zehai He and Zhe Su and Zhen Yang and Ziyang Pan and Aohan Zeng and Baoxu Wang and Bin Chen and Boyan Shi and Changyu Pang and Chenhui Zhang and Da Yin and Fan Yang and Guoqing Chen and Jiazheng Xu and Jiale Zhu and Jiali Chen and Jing Chen and Jinhao Chen and Jinghao Lin and Jinjiang Wang and Junjie Chen and Leqi Lei and Letian Gong and Leyi Pan and Mingdao Liu and Mingde Xu and Mingzhi Zhang and Qinkai Zheng and Sheng Yang and Shi Zhong and Shiyu Huang and Shuyuan Zhao and Siyan Xue and Shangqin Tu and Shengbiao Meng and Tianshu Zhang and Tianwei Luo and Tianxiang Hao and Tianyu Tong and Wenkai Li and Wei Jia and Xiao Liu and Xiaohan Zhang and Xin Lyu and Xinyue Fan and Xuancheng Huang and Yanling Wang and Yadong Xue and Yanfeng Wang and Yanzi Wang and Yifan An and Yifan Du and Yiming Shi and Yiheng Huang and Yilin Niu and Yuan Wang and Yuanchang Yue and Yuchen Li and Yutao Zhang and Yuting Wang and Yu Wang and Yuxuan Zhang and Zhao Xue and Zhenyu Hou and Zhengxiao Du and Zihan Wang and Peng Zhang and Debing Liu and Bin Xu and Juanzi Li and Minlie Huang and Yuxiao Dong and Jie Tang},
      year={2025},
      eprint={2507.01006},
      archivePrefix={arXiv},
      primaryClass={cs.CV},
      url={https://arxiv.org/abs/2507.01006}, 
}

@article{hurst2024gpt,
  title={Gpt-4o system card},
  author={Hurst, Aaron and Lerer, Adam and Goucher, Adam P and Perelman, Adam and Ramesh, Aditya and Clark, Aidan and Ostrow, AJ and Welihinda, Akila and Hayes, Alan and Radford, Alec and others},
  journal={arXiv preprint arXiv:2410.21276},
  year={2024}
}

@article{guo2025seed1,
  title={Seed1. 5-vl technical report},
  author={Guo, Dong and Wu, Faming and Zhu, Feida and Leng, Fuxing and Shi, Guang and Chen, Haobin and Fan, Haoqi and Wang, Jian and Jiang, Jianyu and Wang, Jiawei and others},
  journal={arXiv preprint arXiv:2505.07062},
  year={2025}
}

@techreport{grok4-model-card-2025,
  author       = {{xAI}},
  title        = {Grok 4 Model Card},
  institution  = {xAI},
  year         = {2025},
  month        = {August},
  url          = {https://data.x.ai/2025-08-20-grok-4-model-card.pdf}
}

@misc{seed16vision-2025,
  author       = {{ByteDance Seed Team}},
  title        = {Introduction to Techniques Used in Seed1.6},
  year         = {2025},
  month        = {June},
  howpublished = {\url{https://seed.bytedance.com/en/blog/introduction-to-techniques-used-in-seed1-6}},
  note         = {Official overview; closest public source for Seed 1.6 Vision},
}

@techreport{gpt5-system-card-2025,
  author       = {{OpenAI}},
  title        = {GPT-5 System Card},
  institution  = {OpenAI},
  year         = {2025},
  month        = {August},
  url          = {https://cdn.openai.com/gpt-5-system-card.pdf}
}

@misc{claude-haiku-45-2025,
  author       = {{Anthropic}},
  title        = {Introducing Claude Haiku 4.5},
  year         = {2025},
  month        = {October},
  howpublished = {\url{https://www.anthropic.com/news/claude-haiku-4-5}}
}

@techreport{claude-sonnet-45-system-card-2025,
  author       = {{Anthropic}},
  title        = {Claude Sonnet 4.5 System Card},
  institution  = {Anthropic},
  year         = {2025},
  month        = {October},
  url          = {https://www.anthropic.com/claude-sonnet-4-5-system-card}
}

@misc{claude-opus-41-2025,
  author       = {{Anthropic}},
  title        = {Claude Opus 4.1},
  year         = {2025},
  month        = {August},
  howpublished = {\url{https://www.anthropic.com/news/claude-opus-4-1}}
}

@misc{mistral-medium-3-announce-2025,
  author       = {{Mistral AI}},
  title        = {Medium is the new large: Introducing Mistral Medium 3},
  year         = {2025},
  month        = may,
  howpublished = {\url{https://mistral.ai/news/mistral-medium-3}},
  note         = {Official announcement post}
}

@misc{brendan_adimagenet,
  author       = {Peter Brendan},
  title        = {AdImageNet},
  year         = {2024},
  howpublished = {Hugging Face dataset repository},
  url          = {https://huggingface.co/datasets/PeterBrendan/AdImageNet},
}

@article{ji2025medomni,
  title={Medomni-45 $\{$$\backslash$deg$\}$: A safety-performance benchmark for reasoning-oriented llms in medicine},
  author={Ji, Kaiyuan and Guo, Yijin and Zhang, Zicheng and Zhu, Xiangyang and Tian, Yuan and Liu, Ning and Zhai, Guangtao},
  journal={arXiv preprint arXiv:2508.16213},
  year={2025}
}

@inproceedings{ji2026medomni45,
  title={{MedOmni-45°: A Safety--Performance Benchmark for Reasoning-Oriented LLMs in Medicine}},
  author={Ji, Kaiyuan and Guo, Yijin and Zhang, Zicheng and Zhu, Xiangyang and Tian, Yuan and Liu, Ning},
  booktitle={Proceedings of the AAAI Conference on Artificial Intelligence},
  volume={40},
  number={42},
  pages={35536--35544},
  year={2026},
  doi={10.1609/aaai.v40i42.40864},
  url={https://ojs.aaai.org/index.php/AAAI/article/view/40864}
}

@article{ji2025evaluating,
  title={Evaluating ChatGPT-4's performance on oral and maxillofacial queries: Chain of Thought and standard method},
  author={Ji, Kaiyuan and Wu, Zhihan and Han, Jing and Zhai, Guangtao and Liu, Jiannan},
  journal={Frontiers in Oral Health},
  volume={6},
  pages={1541976},
  year={2025}
}

@article{ji2025assessing,
  title={Assessing the Capabilities of Generative Pretrained Transformer-4 in Addressing Open-Ended Inquiries of Oral Cancer},
  author={Ji, Kaiyuan and Han, Jing and Zhai, Guangtao and Liu, Jiannan},
  journal={International Dental Journal},
  volume={75},
  number={1},
  pages={158--165},
  year={2025},
  publisher={Elsevier}
}

@incollection{ji2024application,
  title={Application of 3D nnU-Net with residual encoder in the 2024 MICCAI head and neck tumor segmentation challenge},
  author={Ji, Kaiyuan and Wu, Zhihan and Han, Jing and Jia, Jun and Zhai, Guangtao and Liu, Jiannan},
  booktitle={Challenge on Head and Neck Tumor Segmentation for MRI-Guided Applications},
  pages={250--258},
  year={2024},
  publisher={Springer}
}

@article{wang2026dental,
  title={Dental-QAD: Reasoning-driven quality assessment and diagnosis in panoramic radiographs},
  author={Wang, Shuangqing and Ji, Kaiyuan and Zheng, Yushuo and Wu, Zhihan and Zhu, Xiaorong and Chen, Zijian and Sun, Lu and Wang, Shuo and Zhang, Jianbo and Zhang, Zicheng and others},
  journal={Displays},
  pages={103380},
  year={2026},
  publisher={Elsevier}
}

@article{lei2026sequential,
  title={Sequential sensitivity analysis of multimodal large language models for rare orbital disease detection},
  author={Lei, Chaoyu and Ji, Kaiyuan and Zhao, Chen and Zhong, Sisi and Cao, Chenyu and Chen, Hao and Yip, Chee Chew and Sintuwong, Sunisa and Ding, Jianbin and Pandiyan, PS and others},
  journal={Communications Medicine},
  year={2026},
  publisher={Nature Publishing Group UK London}
}

@article{liang2025priceseer,
  title={PriceSeer: Evaluating Large Language Models in Real-Time Stock Prediction},
  author={Liang, Bohan and Chen, Zijian and Jia, Qi and Zhang, Kaiwei and Ji, Kaiyuan and Zhai, Guangtao},
  journal={arXiv preprint arXiv:2601.06088},
  year={2025}
}

@article{guo2025human,
  title={Human-centric evaluation for foundation models},
  author={Guo, Yijin and Ji, Kaiyuan and Zhu, Xiaorong and Wang, Junying and Wen, Farong and Li, Chunyi and Zhang, Zicheng and Zhai, Guangtao},
  journal={arXiv preprint arXiv:2506.01793},
  year={2025}
}

@article{jin2025medscreendental,
  title={MedScreenDental: Automated structured dental record generation via multimodal language model integration},
  author={Jin, Wenzhong and Sun, Yilan and Ji, Kaiyuan and Jiang, Xiaoyan and Hu, Yufeng and Wang, Jinwu and Liu, Jiannan},
  journal={Displays},
  volume={90},
  pages={103119},
  year={2025},
  publisher={Elsevier}
}

@article{wang2025quality,
  title={Quality evaluation of large language models in answering open-ended Questions in the field of Benign prostatic hyperplasia},
  author={Wang, Chengbang and Liu, Zijia and Hao, Liangshi and Chen, Shaohua and Guo, Rongchang and Wei, Lai and Ji, Kaiyuan and Wang, Fubo and Xu, Bin},
  journal={Displays},
  volume={90},
  pages={103144},
  year={2025},
  publisher={Elsevier}
}

@article{tian2025smc++,
  title={Smc++: Masked learning of unsupervised video semantic compression},
  author={Tian, Yuan and Ling, Xiaoyue and Geng, Cong and Hu, Qiang and Lu, Guo and Zhai, Guangtao},
  journal={IEEE Transactions on Pattern Analysis and Machine Intelligence},
  year={2025},
  publisher={IEEE}
}

@article{tian2024coding,
  title={A coding framework and benchmark towards low-bitrate video understanding},
  author={Tian, Yuan and Lu, Guo and Yan, Yichao and Zhai, Guangtao and Chen, Li and Gao, Zhiyong},
  journal={IEEE Transactions on Pattern Analysis and Machine Intelligence},
  volume={46},
  number={8},
  pages={5852--5872},
  year={2024},
  publisher={IEEE}
}

@article{tian2025rofi,
  title={ROFI: A Deep Learning-Based Ophthalmic Sign-Preserving and Reversible Patient Face Anonymizer},
  author={Tian, Yuan and Zhou, Min and Chen, Yitong and others},
   journal={npj Digital Medicine},
  year={2025},
  publisher={Nature Publishing Group UK London}
}

@inproceedings{tian2025towards,
  title={Towards All-in-One Medical Image Re-Identification},
  author={Tian, Yuan and Ji, Kaiyuan and Zhang, Rongzhao and Jiang, Yankai and Li, Chunyi and Wang, Xiaosong and Zhai, Guangtao},
  booktitle={Proceedings of the Computer Vision and Pattern Recognition Conference},
  pages={30774--30786},
  year={2025}
}

@inproceedings{tian2025semantic,
  title={Semantic versus Identity: A Divide-and-Conquer Approach towards Adjustable Medical Image De-Identification},
  author={Tian, Yuan and Wang, Shuo and Zhang, Rongzhao and others},
  booktitle={Proceedings of the IEEE/CVF International Conference on Computer Vision},
  pages={20613--20625},
  year={2025}
}

@article{gao2025blind,
  title={Blind image quality assessment by Gaussian mixture distribution},
  author={Gao, Yixuan and Min, Xiongkuo and Cao, Yuqin and Lin, Weisi and Lee, Bu Sung and Zhai, Guangtao},
  journal={IEEE Transactions on Image Processing},
  year={2025},
  publisher={IEEE}
}

@inproceedings{gao2025multi,
  title={Multi-Dimensional Text-to-Face Image Quality Assessment Using LLM: Database and Method},
  author={Gao, Yixuan and Min, Xiongkuo and Han, Jinliang and Cao, Yuqin and Wu, Sijing and Dou, Yunze and Zhai, Guangtao},
  booktitle={Proceedings of the 33rd ACM International Conference on Multimedia},
  pages={6948--6957},
  year={2025}
}

@article{gao2024no,
  title={No-reference image quality assessment: Obtain mos from image quality score distribution},
  author={Gao, Yixuan and Min, Xiongkuo and Cao, Yuqin and Liu, Xiaohong and Zhai, Guangtao},
  journal={IEEE Transactions on Circuits and Systems for Video Technology},
  volume={35},
  number={2},
  pages={1840--1854},
  year={2024},
  publisher={IEEE}
}

@article{chen2025just,
  title={Just Noticeable Difference for Large Multimodal Models},
  author={Chen, Zijian and Tian, Yuan and Sun, Yuze and Sun, Wei and Zhang, Zicheng and Lin, Weisi and Zhai, Guangtao and Zhang, Wenjun},
  journal={arXiv preprint arXiv:2507.00490},
  year={2025}
}

@article{si2024accelerating,
  title={Accelerating non-maximum suppression: a graph theory perspective},
  author={Si, King-Siong and Sun, Lu and Zhang, Weizhan and Gong, Tieliang and Wang, Jiahao and Liu, Jiang and Sun, Hao},
  journal={Advances in Neural Information Processing Systems},
  volume={37},
  pages={121992--122028},
  year={2024}
}

@misc{li2025quantumapproximateoptimizationalgorithms,
      title={Quantum Approximate Optimization Algorithms for Maximum Cut on Low-Girth Graphs}, 
      author={Tongyang Li and Yuexin Su and Ziyi Yang and Shengyu Zhang},
      year={2025},
      eprint={2410.04409},
      archivePrefix={arXiv},
      primaryClass={quant-ph},
      url={https://arxiv.org/abs/2410.04409}, 
}

@article{li2025perceptual,
  title={Perceptual quality assessment for embodied ai},
  author={Li, Chunyi and Xiao, Jiaohao and Zhang, Jianbo and Wen, Farong and Zhang, Zicheng and Tian, Yuan and Zhu, Xiangyang and Liu, Xiaohong and Cheng, Zhengxue and Lin, Weisi and others},
  journal={arXiv e-prints},
  pages={arXiv--2505},
  year={2025}
}

@article{cao2025towards,
  title={Towards Generalized Video Quality Assessment: A Weak-to-Strong Learning Paradigm},
  author={Cao, Linhan and Sun, Wei and Zhu, Xiangyang and Zhang, Kaiwei and Jia, Jun and Peng, Yicong and Zhu, Dandan and Zhai, Guangtao and Min, Xiongkuo},
  journal={arXiv preprint arXiv:2505.03631},
  year={2025}
}

@article{jiang2026surveillance,
  title={Surveillance Facial Image Quality Assessment: A Multi-dimensional Dataset and Lightweight Model},
  author={Jiang, Yanwei and Sun, Wei and Zhou, Yingjie and Zhu, Xiangyang and Cao, Yuqin and Jia, Jun and Li, Yunhao and Wu, Sijing and Zhu, Dandan and Min, Xingkuo and others},
  journal={IEEE Transactions on Circuits and Systems for Video Technology},
  year={2026},
  publisher={IEEE}
}

@misc{zheng2026learningwanderimprovingglobal,
      title={Learning to Wander: Improving the Global Image Geolocation Ability of LMMs via Actionable Reasoning}, 
      author={Yushuo Zheng and Huiyu Duan and Zicheng Zhang and Xiaohong Liu and Xiongkuo Min},
      year={2026},
      eprint={2603.10463},
      archivePrefix={arXiv},
      primaryClass={cs.CV},
      url={https://arxiv.org/abs/2603.10463}, 
}

@misc{zheng2025lmfightarenabenchmarking,
      title={LM Fight Arena: Benchmarking Large Multimodal Models via Game Competition}, 
      author={Yushuo Zheng and Zicheng Zhang and Xiongkuo Min and Huiyu Duan and Guangtao Zhai},
      year={2025},
      eprint={2510.08928},
      archivePrefix={arXiv},
      primaryClass={cs.AI},
      url={https://arxiv.org/abs/2510.08928}, 
}

@inproceedings{wang2025learning,
    title={Learning Hazing to Dehazing: Towards Realistic Haze Generation for Real-World Image
  Dehazing},
    author={Wang, Ruiyi and Zheng, Yushuo and Zhang, Zicheng and Li, Chunyi and Liu, Shuaicheng and
  Zhai, Guangtao and Liu, Xiaohong},
    booktitle={Proceedings of the IEEE/CVF Conference on Computer Vision and Pattern Recognition},
    year={2025}
  }

@inproceedings{jin2025rgcvqa,
    title={RGC-VQA: An Exploration Database for Robotic-Generated Video Quality Assessment},
    author={Jin, Jian and Ying, Jiangyong and Duan, Huiyu and Yang, Liu and Wu, Sijing and Li,
  Yunhao and Zheng, Yushuo and Min, Xiongkuo and Zhai, Guangtao},
    booktitle={Proceedings of the ACM International Conference on Multimedia},
    year={2025}
  }

@article{zheng2025geoxbench,
    title={GeoX-Bench: Benchmarking Cross-View Geo-Localization and Pose Estimation Capabilities of
  Large Multimodal Models},
    author={Zheng, Yushuo and Ying, Jiangyong and Duan, Huiyu and Li, Chunyi and Zhang, Zicheng and
  Liu, Jing and Liu, Xiaohong and Zhai, Guangtao},
    journal={arXiv preprint arXiv:2511.13259},
    year={2025}
  }

@article{AIBench,
  title   = {AIBench: Towards trustworthy evaluation under the 45° law},
  author  = {Zicheng Zhang and Junying Wang and Yijin Guo and others},
  journal = {Displays},
  year    = {2025},
  pages   = {103255},
  issn    = {0141-9382},
  doi     = {10.1016/j.displa.2025.103255}
}

@article{zhang2025large,
  author    = {Zhang, Zicheng and Wang, Junying and Wen, Farong and Guo, Yijin and others},
  title     = {Large Multimodal Models Evaluation: A Survey},
  journal   = {SCIENCE CHINA Information Sciences},
  year      = {2025},
  number  = {12},
  volume    = {68},
  pages     = {221301-221369},
  doi       = {https://doi.org/10.1007/s11432-025-4676-4}
}

@article{zhang2025teaching,
  title={Teaching lmms for image quality scoring and interpreting},
  author={Zhang, Zicheng and Wu, Haoning and Jia, Ziheng and Lin, Weisi and Zhai, Guangtao},
  journal={arXiv preprint arXiv:2503.09197},
  year={2025}
}

@article{zhang2025quality,
  title={Quality assessment in the era of large models: A survey},
  author={Zhang, Zicheng and Zhou, Yingjie and Li, Chunyi and Zhao, Baixuan and Liu, Xiaohong and Zhai, Guangtao},
  journal={ACM Transactions on Multimedia Computing, Communications and Applications},
  volume={21},
  number={7},
  pages={1--31},
  year={2025},
  publisher={ACM New York, NY}
}

@article{zhang2022no,
  title={No-reference quality assessment for 3D colored point cloud and mesh models},
  author={Zhang, Zicheng and Sun, Wei and Min, Xiongkuo and Wang, Tao and Lu, Wei and Zhai, Guangtao},
  journal={IEEE Transactions on Circuits and Systems for Video Technology},
  volume={32},
  number={11},
  pages={7618--7631},
  year={2022},
  publisher={IEEE}
}

@inproceedings{wang2025bamnet,
  title={BAMNet: A Brain Area Mapping-Based Multimodal Saliency Prediction Method},
  author={Wang, Shibo},
  booktitle={Proceedings of the IEEE International Conference on Multimedia and Expo (ICME)},
  year={2025},
  pages={1--6},
  doi={10.1109/ICME59968.2025.11209775}
}

@inproceedings{wang2026brain,
  title     = {A Brain-Inspired Saliency Prediction Framework for Human-AI Cognitive Consistency in AIGC Content via Multi-Region Liquid Neurons},
  author    = {Wang, Shibo and Zhao, Yan and Wang, Shigang and Wei, Jian and Li, Shuo},
  booktitle = {Proceedings of the AAAI Conference on Artificial Intelligence},
  year      = {2026},
  volume    = {40},
  number    = {21},
  url       = {https://ojs.aaai.org/index.php/AAAI/issue/view/702}
}
}

\end{document}